\pdfoutput=1

\documentclass[11pt]{article}

\usepackage[dvipsnames,table]{xcolor}

\usepackage[preprint]{acl}

\usepackage{times}
\usepackage{latexsym}

\usepackage[T1]{fontenc}

\usepackage[utf8]{inputenc}

\usepackage{microtype}

\usepackage{inconsolata}

\usepackage[table]{xcolor}
\usepackage{graphicx}
\usepackage{tabularx}
\usepackage{booktabs}
\usepackage{float}

\usepackage{multirow}
\usepackage{array}
\usepackage{amsmath}
\usepackage{makecell}
\usepackage{enumitem}
\usepackage{placeins}

\usepackage{fontawesome5}

\usepackage[normalem]{ulem}
\usepackage{enumitem}

\newcommand{\taxonomy}[0]{\textsc{LiTEx}}
\newcommand{\dataset}[0]{\textsc{LiTEx-SNLI}}

\title{{\taxonomy}: A \textsc{Li}nguistic \textsc{T}axonomy of \textsc{Ex}planations for Understanding Within-Label Variation in Natural Language Inference}

\author{
\textbf{Pingjun Hong\thanks{\; Equal contribution.}\thanks{\; Main work carried out while at LMU Munich.}\textsuperscript{\faMountain\kern1pt\faLaptopCode}} \quad
\textbf{Beiduo Chen\footnotemark[1]\textsuperscript{\faMountain\kern1pt\faRobot}} \quad
\textbf{Siyao Peng\textsuperscript{\faMountain\kern1pt\faRobot}} \quad
\\
\textbf{Marie-Catherine de Marneffe\textsuperscript{\faPenFancy}} \quad
\textbf{Barbara Plank\textsuperscript{\faMountain\kern1pt\faRobot}}
\\[8pt]
\textsuperscript{\faMountain}MaiNLP, Center for Information and Language Processing, LMU Munich, Germany \\
\textsuperscript{\faRobot}Munich Center for Machine Learning, Germany
\textsuperscript{\faPenFancy}FNRS, CENTAL, UCLouvain, Belgium \\
\textsuperscript{\faLaptopCode}Faculty of Computer Science and UniVie Doctoral School Computer Science, \\University of Vienna, Austria \\
{\tt{ \href{mailto:pingjun.hong@univie.ac.at}{\textcolor{black}{pingjun.hong@univie.ac.at}}, \{\href{mailto:beiduo.chen@lmu.de}{\textcolor{black}{beiduo.chen}}, \href{mailto:siyao.peng@lmu.de}{\textcolor{black}{siyao.peng}}, \href{mailto:b.plank@lmu.de}{\textcolor{black}{b.plank}}\}@lmu.de}}, \\ {\tt{\href{mailto:marie-catherine.demarneffe@uclouvain.be}{\textcolor{black}{marie-catherine.demarneffe@uclouvain.be}}}}}

\begin{document}
\maketitle
\begin{abstract}
There is increasing evidence of Human Label Variation (HLV) in Natural Language Inference (NLI), where annotators assign different labels to the same premise-hypothesis pair.
However, \textit{within-label variation}---cases where annotators agree on the same label but provide divergent reasoning---poses an additional and mostly overlooked challenge.
Several NLI datasets contain highlighted words in the NLI item as explanations, but the same spans on the NLI item can be highlighted for different reasons, as evidenced by free-text explanations, which offer a window into annotators' reasoning.
To systematically understand this problem and gain insight into the rationales behind NLI labels, we introduce \taxonomy, a linguistically-informed taxonomy for categorizing free-text explanations in English.
Using this taxonomy, we annotate a subset of the e-SNLI dataset, validate the taxonomy's reliability, and analyze how it aligns with NLI labels, highlights, and explanations.
We further assess the taxonomy's 
role in explanation generation, demonstrating that conditioning generation on \taxonomy{} yields explanations that are linguistically closer to human explanations than those generated using only labels or highlights. Our approach thus not only captures within-label variation but also shows how taxonomy-guided generation for reasoning can bridge the gap between human and model explanations more effectively than existing strategies.
\end{abstract}

\begin{figure}[t]
  \includegraphics[width=\columnwidth]{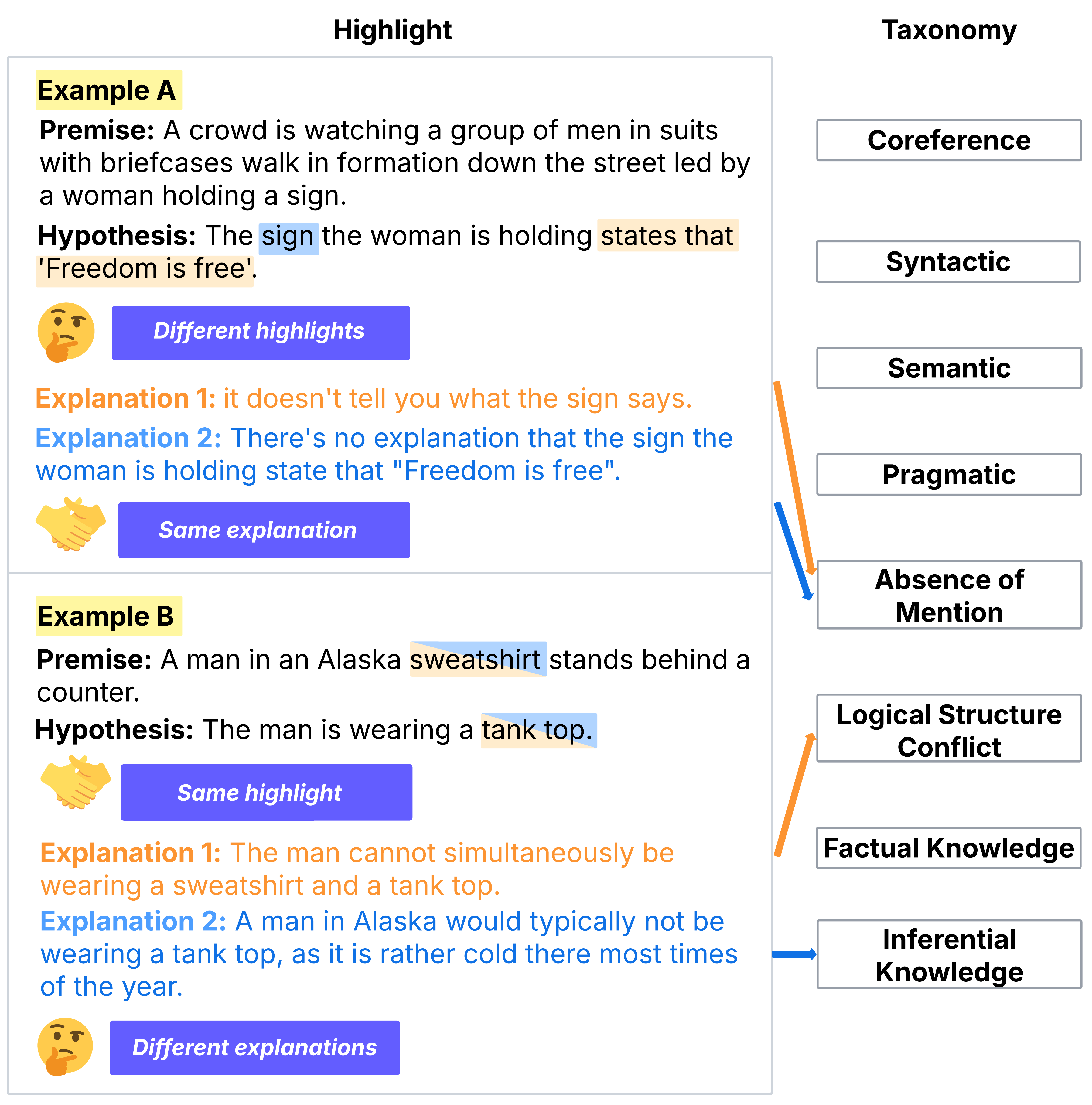}
  \caption{Our \taxonomy{} taxonomy reveals within-label variation not captured by highlights: the same highlights can yield different explanations (Example B), and vice versa (Example A). 
  }
  \label{fig:example}
\end{figure}

\section{Introduction}

Natural Language Inference (NLI), a cornerstone task in Natural Language Processing (NLP), has inspired extensive research on human disagreement and model interpretability. A key focus of recent work has been Human Label Variation (HLV, \citealt{plank-2022-problem}) — cases in which annotators assign different labels to the same premise-hypothesis pair \citep{nie-etal-2020-learn, jiang_ecologically_2023, weber-genzel_varierr_2024}. This variation has been acknowledged as a reflection of subjective judgment \cite{Cabitza_2023} and linguistic ambiguity \cite{de-marneffe-etal-2012-happen, Uma/jair.1.12752}. Comparatively, the issue of \textit{within-label variation} \citep{jiang_ecologically_2023} -- cases where annotators agree on the same label, yet provide different explanations or rationales for their decision -- has received less attention. Such variation reveals the plurality of valid reasoning strategies and highlights the richness of human inference beyond label selection.

Free-text explanations offer a rich perspective on reasoning variation. However, their open-ended form makes it difficult to extract information that is directly useful for downstream analysis. As a result, structured formats are often used when collecting human explanations. Highlights are one such mechanism \cite{tan-2022-diversity}. 
\citet{jiang_ecologically_2023} acknowledge that textual highlight spans alone are insufficient to capture deeper reasoning distinctions including within-label variation, especially when explanations focus on different parts of the input or rely on different assumptions. As illustrated in Figure~\ref{fig:example}, two explanations in Example B may share the same highlighted spans (here \textit{sweatshirt} and \textit{tank top}) but reflect different reasoning strategies (one annotator focuses on the fact that sweatshirt and tank top are not typically worn together, whereas the other says that one does not wear a tank top in Alaska); or conversely, different highlights may convey essentially the same explanation, as seen in Example A. 

To address this gap, \textbf{(1)} we introduce \taxonomy{}, a \textsc{Li}nguistic \textsc{T}axonomy of \textsc{Ex}planations for understanding within-label variation in English natural language inference explanations. \textbf{(2)} We validate our taxonomy through human inter-annotator agreement and model-based classification. We further analyze its alignment with NLI labels and quantify within-label variation by examining category distribution and their similarity—demonstrating the taxonomy’s ability to capture different types of explanations. \textbf{(3)} While human explanations are costly, LLMs offer a scalable alternative for generating explanations in NLI \cite{chen-etal-2025-rose}. Through generation experiments, we demonstrate that taxonomy-based guidance provides a more effective signal for LLMs than highlight-based prompts.

\section{Related Work}

\paragraph{Explaining NLI Labels}
Explanations play a crucial role in making NLI decisions interpretable. As \citet{tan-2022-diversity} highlights, explanations vary in form and quality, and improving their usefulness requires distinguishing between different explanation types and recognizing human limitations in producing them. Among existing methods, token-level highlights serve as a proxy for explanations, guiding annotators to mark relevant spans that support their label choice. Several NLI datasets provide such annotations (including free-text explanations also), collected either during labeling (e.g., LiveNLI \cite{jiang_ecologically_2023} and ANLI \cite{nie-etal-2020-adversarial}) or post-hoc (e.g., e-SNLI \cite{camburu2018esnlinaturallanguageinference}). Here, we focus on both types of explanations (free-text and highlights) from e-SNLI.

\paragraph{Taxonomies of Variation in NLI}

In the context of NLI, earlier taxonomies focused on categorizing the kind of inferences present in NLI items \cite{sammons-etal-2010-ask, article, lobue-yates-2011-types}. Later work proposed a taxonomy that identifies characteristics of the items that can cause variation in annotation \cite{jiang-marneffe-2022-investigating}. \citet{jiang_ecologically_2023} shifted the focus from the NLI items, collecting free-text explanations provided by the annotators themselves, applying \citet{jiang-marneffe-2022-investigating}'s taxonomy to the explanations. \citet{jayaweera2025disagreementunderstandingcaseambiguity} further argued for an ambiguity-aware NLI framework that detects ambiguous instances and classifies them using the taxonomy of \citet{jiang-marneffe-2022-investigating}.  

Our work builds on this direction by proposing a taxonomy of explanations for instances that share the same NLI label, aiming to capture within-label variation in reasoning. Compared to \citet{jiang_ecologically_2023}, our taxonomy is thus grounded in the explanations. It also makes world knowledge in NLI reasoning explicit.

\paragraph{LLM-Based Explanation Generation}
Recent studies explored the use of LLMs to generate natural language explanations across a range of 
NLP tasks, aiming to improve transparency and support downstream analysis. 
\citet{li2022explanationslargelanguagemodels} proposed prompting LLMs to generate chain-of-thought (CoT) explanations to improve the performance of small task-specific models. \citet{chen2025threadingneedlereweavingchainofthought} further repurposed CoTs as a forward source of explanation-label pairs, applying discourse-guided segmentation to extract structured rationales. \citet{huang2023largelanguagemodelsexplain} investigated whether LLMs could generate faithful self-explanations to justify their own predictions during inference.

In NLI, \citet{jiang_ecologically_2023} employed GPT-3 to generate post-prediction explanations (predict-then-explain) and found this strategy to outperform CoT prompting.  \citet{chen-etal-2025-rose} showed that LLMs can effectively generate explanations to approximate human judgment distribution, offering a scalable and cost-efficient alternative to manual annotation. Building on this line of work, we use our proposed taxonomy to guide LLM prompting for more informative and human-aligned explanations.

\begin{table*}[t!bh]
  \centering
  \footnotesize
  \renewcommand{\arraystretch}{0.98}
  \resizebox{\textwidth}{!}{%
  \begin{tabularx}{\textwidth}{clX}
    \toprule
 \multicolumn{3}{c}{\textbf{Text-Based Reasoning (TB)}} \\ 
 \midrule
    \multirow{2}{*}{\textit{Coreference 
    }} &
    \textbf{Q:} & \textit{Does the explanation rely on resolving coreference between entities?} \\
   & \textbf{Check:} & Determine whether the main entities in the premise and hypothesis refer to the same real-world referent, including via pronouns or phrases.\\
   \midrule
   \multirow{2}{*}{\textit{Syntactic}} & \textbf{Q:} & \textit{Does the explanation involve a change in sentence structure that preserves meaning?} \\
   &    \textbf{Check:} & Determine whether the premise and hypothesis differ in structure, such as active vs. passive, reordered arguments, or coordination/subordination, while preserving the same meaning.\\
   \midrule
   \multirow{2}{*}{\textit{Semantic}} & \textbf{Q:} & \textit{Does the explanation involve semantic similarity or substitution of key concepts?} \\
    &   \textbf{Check:} & Evaluate whether core words or expressions - including verbs, nouns, and adjectives - are semantically related between the premise and hypothesis. This includes synonymy, antonymy, lexical entailment, or category membership.\\
    \midrule
   \multirow{2}{*}{\textit{Pragmatic}} & \textbf{Q:} & \textit{Does the explanation rely on pragmatic cues like implicature or presupposition?} \\
   &    \textbf{Check:} & Look for meaning beyond the literal text - including implicature, presupposition, speaker intention, and conventional conversational meaning.\\
   \midrule
   \multirow{2}{*}{\begin{tabular}[c]{@{}c@{}}\textit{Absence} \\ \textit{ of Mention}\end{tabular}} & \textbf{Q:} & \textit{Does the explanation point out information not mentioned in the premise?} \\
   &    \textbf{Check:} & Check whether the hypothesis introduced information that is neither supported nor contradicted by the premise - i.e., it is not mentioned explicitly.\\
   \midrule
 \multirow{2}{*}{\begin{tabular}[c]{@{}c@{}}\textit{Logic} \\ \textit{Conflict}\end{tabular}} &   \textbf{Q:} & \textit{Does the explanation refer to logical constraints or conflict?} \\
   &    \textbf{Check:} & Evaluate whether the hypothesis interacts with the premise via logical structures, such as exclusivity, quantifiers (“only”, “none”), or conditionals, which constrain or conflict with each other.\\ \midrule
    \multicolumn{3}{c}{\textbf{World Knowledge-Based Reasoning (WK)}} \\
    \midrule
 \multirow{2}{*}{
 \begin{tabular}[c]{@{}c@{}}
 \textit{Factual} \\ 
 \textit{Knowledge} \end{tabular}
 } &   \textbf{Q:} & \textit{Does the explanation rely on widely shared, intuitive facts acquired through everyday experience?} \\
   &    \textbf{Check:} & Determine whether the explanation invokes commonly known facts, such as physical properties or universal experiences, that are not stated in the premise. \\
   \midrule
  \multirow{2}{*}{
  \begin{tabular}[c]{@{}c@{}}
  \textit{Inferential} \\
  \textit{Knowledge}
  \end{tabular}
  } &  \textbf{Q:} & \textit{Does the explanation rely on real-world norms, customs, or culturally grounded reasoning?} \\
 &   \textbf{Check:} & Determine whether the explanation requires reasoning based on general world knowledge, including cultural expectations, social norms, or typical causal inferences, that are not stated in the premise.\\ \bottomrule
  \end{tabularx}}
  \caption{Guiding questions and decision criteria for our \taxonomy{} taxonomy.}
  \label{tab:taxonomy_questions}
\end{table*}

\section{\taxonomy: Linguistically-informed Taxonomy of NLI Reasoning}
\label{sec:taxonomy}
To systematically capture the different types of reasoning strategies underlying within-label variation in NLI, we propose \taxonomy, a \textsc{Li}nguistic Taxonomy of 
\textsc{Ex}planation classification, focusing strictly on the reasoning explicitly stated in the explanations. 

\subsection{Taxonomy Categories}
\taxonomy{} organizes explanations into two broad categories based on their reliance on textual evidence or external knowledge, as shown in Table~\ref{tab:taxonomy_questions}. This categorization builds on the work of \citet{jiang-marneffe-2022-investigating}.

The first broad category, \textit{Text-Based (TB) Reasoning}, includes explanations that depend solely on \textit{surface-level linguistic evidence found within the premise and hypothesis}, without appealing to world knowledge. 
Six subtypes are defined: \textit{Coreference}, \textit{Syntactic}, \textit{Semantic}, \textit{Pragmatic}, \textit{Absence of Mention} and \textit{Logic Conflict}. 

The second category, \textit{World-Knowledge (WK) Reasoning}, includes explanations that invoke background knowledge or domain-specific information beyond what is explicitly stated in the text. \textit{Factual knowledge} refers to widely shared, intuitive facts acquired through everyday experience, such as \textit{fire is hot}. \textit{Inferential Knowledge} involves culturally or contextually grounded understanding, such as recognizing that \textit{wearing white to a funeral is inappropriate} (a norm that varies across cultures) \cite{10.5555/3176788.3176804, ilievski2021dimensionscommonsenseknowledge}. 

Table~\ref{tab:taxonomy_questions} presents guiding questions and decision criteria for each taxonomy category to help annotators identify the reasoning behind explanations. 
These questions, along with illustrative examples in Appendix \ref{sec:Taxonomy_examples}, clarify the conceptual boundaries between categories. For example, to distinguish between \textit{Logic Conflict} and \textit{Semantic}, consider the following two explanations: (a) \textit{A man cannot be both tall and short at the same time} and (b) \textit{Tall and short are not the same}.
Explanation (a) reflects a logical inconsistency, pointing to the mutual exclusivity of properties, and thus labeled as \textit{Logic Conflict}, whereas explanation (b) highlights lexical contrast or antonymy without explicit logical reasoning, and thus \textit{Semantic}.

\subsection{Taxonomy Annotation}\label{subsec:taxonomy-annotation}
We randomly selected a subset (1,002 items) of the e-SNLI dataset, in which each item received three post-hoc human-written explanations accompanied by highlights. 
We conduct \taxonomy{} annotations on these explanations. 
To better capture distinct reasoning strategies, we manually segment the long explanations that potentially include multiple inferences into shorter ones. As a result, the original 3,006 explanations are expanded to 3,108. One trained annotator applied \taxonomy{} to these 3,108 explanations (and the associated premise, hypothesis, and NLI label are provided as context), labeling each with one of the eight categories. 

\subsection{Taxonomy Validation}

To validate the consistency and generalizability of our \taxonomy{} taxonomy, we provide human inter-annotator agreement (IAA) and benchmark experiments on automatic explanation classification. 

\paragraph{IAA}\label{subsec:iaa}
We assess the consistency of our human annotations by calculating IAA on a subset of the e-SNLI explanations, separate from \dataset{} used in our main experiments. Two annotators, the one from the initial phase and one newly recruited,\!\footnote{Both are trained and paid according to national standards.}
annotated 201 explanations from 67 extra e-SNLI items, using the proposed taxonomy. The agreement is high (Cohen's $k$ of 0.862), suggesting that the taxonomy can be applied consistently between annotators. 
Appendix~\ref{sec:IAA-report} presents the confusion matrix and representative examples of annotation disagreements.

\paragraph{Taxonomy Classification}\label{subsec:taxonomy-classification}

\begin{table}[t]
\centering
\footnotesize
\resizebox{\linewidth}{!}{%
\begin{tabular}{lrrrr}
\toprule
\textbf{Classifiers} & \textbf{Acc} & \textbf{P
} & \textbf{R
} & \textbf{F1
} 
\\
\midrule
\textbf{Random Baseline} & 12.5 & 11.8 & 10.8 & 10.2  \\
\textbf{Majority Baseline} & 31.3 & 3.9 & 12.5 & 6.0 \\
\midrule
\textbf{BERT-base
} & \textbf{70.2} & \textbf{60.5} & \textbf{57.9} & \textbf{57.8} 
\\
\textbf{RoBERTa-base
}&68.9& 48.4& 53.4& 50.4
\\
\midrule
\textbf{Llama-3.2-3B-Instruct} & 35.7 & 44.0 & 35.7 & 29.1 
\\

\textbf{gpt-3.5-turbo} & 30.5 & 31.7 & 30.5 & 26.2 
\\

\textbf{gpt-4o} & 58.3 & 55.0 & 54.8 & 49.2 
\\

\textbf{DeepSeek-v3} & 52.6 & 51.9 & 56.3 & 47.8 
\\
\bottomrule
\end{tabular}
}
\caption{Taxonomy classification results (\%) on \dataset. Fine-tuning methods are evaluated with a 50/50 data split; Prompt-based methods use taxonomy descriptions with two examples per category. P(recision), R(ecall), and F1 are at the macro-level.}
\label{tab:llm-classification-main}
\end{table}

\begin{figure}[t]
  \includegraphics[width=\columnwidth]{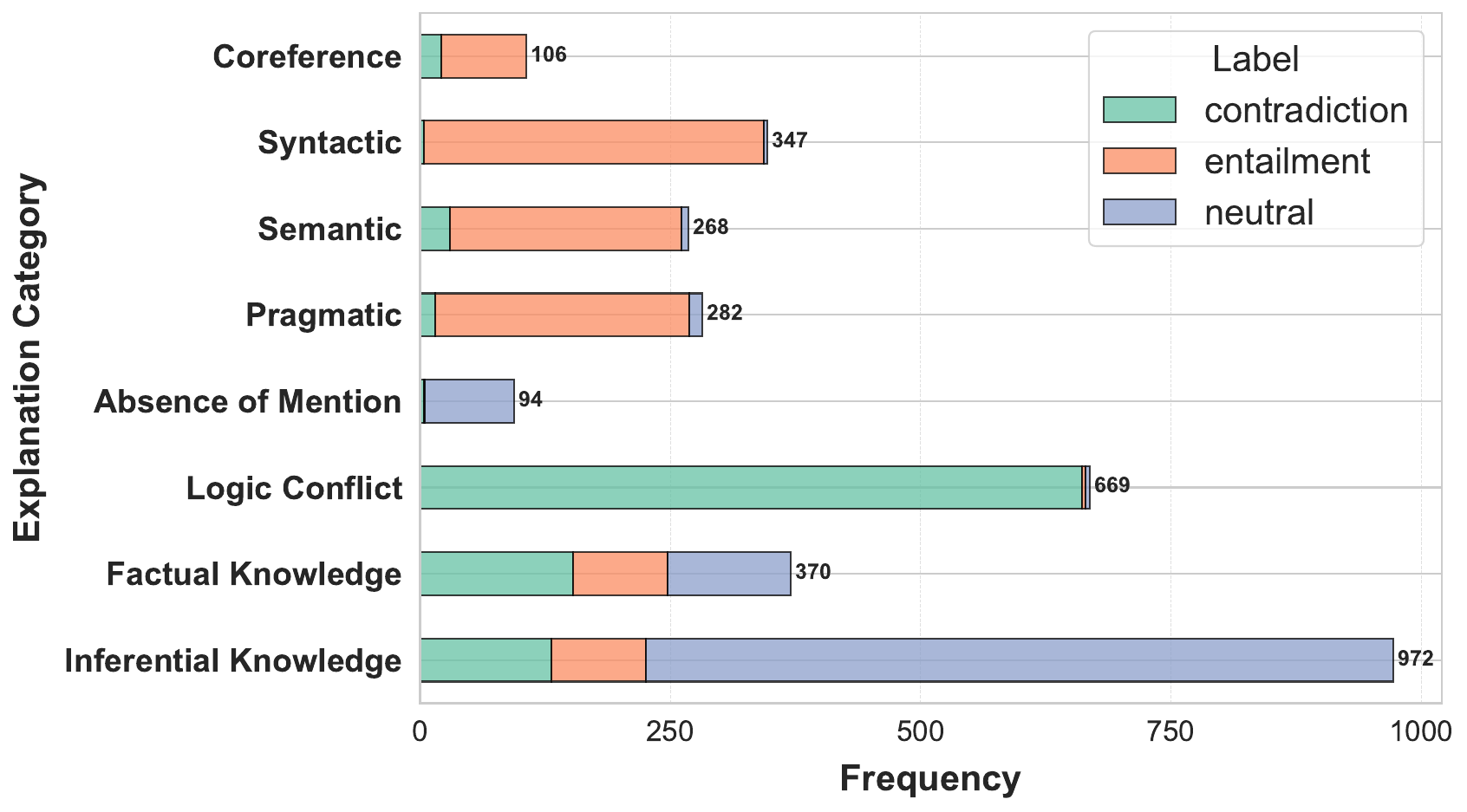}
  \caption{Distribution of \taxonomy{} categories on \dataset{} explanations across NLI labels (n = 3,108).}
  \label{fig:annotation_distribution}
\end{figure}

\begin{table}[t]
\centering
\resizebox{\linewidth}{!}{%
\begin{tabular}{lrrrr}
\toprule
\makecell{\textbf{Category} \\ \textbf{\#}} & \makecell{\textbf{Entailment} \\ \# (\%)} & \makecell{\textbf{Neutral} \\ \# (\%)} & \makecell{\textbf{Contradiction} \\ \# (\%)} & \textbf{Total}\\
\midrule
1 & 76 (22.0)  & 171 (52.3) & 142 (43.0) & 389\\
2 & 179 (51.9) & 139 (42.5)  & 156 (47.3) & 474\\
$\geq3$ & 90 (26.1)  & 17 \; (5.1) & 32  \; (9.7) & 139\\
\bottomrule
\end{tabular}
}
\caption{Distribution of NLI items that receive 1, 2, or >=3  \taxonomy{} categories on their explanations (n = 1,002).}
\label{tab:instance_distribution}
\end{table}

To validate the taxonomy and test its usefulness for automated classification, we fine-tuned two pre-trained language models, BERT-base-uncased \cite{devlin_bert_2019} and RoBERTa-base \cite{liu_roberta_2019}, to classify explanations in \dataset{} to the annotated \taxonomy{} categories.
We also few-shot prompt 4 generative AI models: Llama-3.2-3B-Instruct \cite{llama_3}, GPT-3.5-turbo \cite{brown_language_2020}, GPT-4o \cite{openai_gpt-4_2024} and DeepSeek-v3 \cite{deepseekai2025deepseekv3technicalreport}; see Appendix~\ref{sec:classification-details} for details.

Table~\ref{tab:llm-classification-main} gives the classification results. BERT-base and RoBERTa-base achieve strong results on this 8-way classification task, with macro-F1 scores of 57.8\% and 50.4\%, and accuracies of 70.2\% and 68.9\%, respectively. These results substantially surpass both a random baseline of 12.5\% and a majority-class baseline of 31.3\% (based on the dominant category, \textit{Inferential Knowledge}), emphasizing the benefits of task-specific supervision. 
LLMs, when prompted with detailed taxonomy descriptions and illustrative examples, also perform better than random and majority-class baselines, further confirming our taxonomy's learnability. 

In sum, the findings suggest that the proposed taxonomy is learnable, reinforcing its applicability for both annotation and LLM-based reasoning. 

\subsection{Taxonomy Analysis}
\paragraph{Co-occurrence of Explanation Categories and NLI Labels}

Figure~\ref{fig:annotation_distribution} plots the distribution of our explanation categories and their co-occurrence with NLI labels. We observe that different explanation categories show distinct distributions over NLI labels. \textit{Logic Conflict} is dominated by contradiction, because this category focuses on capturing logical inconsistency. \textit{Syntactic}, \textit{Semantic}, and \textit{Pragmatic} are primarily associated with entailment, suggesting that these reasoning types tend to support alignment. 
\textit{Factual Knowledge} and \textit{Inferential Knowledge} are more evenly distributed across the labels, since world knowledge could be involved in different inference scenarios. Lastly, \textit{Absence of Mention} aligns strongly with neutral, consistent with its reliance on unstated information.  

\paragraph{Within-label Variation }\label{subsec:variation-taxonomy-explanation}

Table~\ref{tab:instance_distribution} gives the counts of our 1,002 NLI items for which the three (or more) explanations were annotated with 1, 2, or $\geq3$ \taxonomy{} categories (cf. \S\ref{subsec:taxonomy-annotation} for explanation segmentation). 
These counts show that within-label variation is prevalent in e-SNLI, e.g., 613 out of 1,002 (61.2\%) items received more than one taxonomy category across explanations. 

\begin{figure}[t]
    \centering
    \includegraphics[width=\columnwidth]{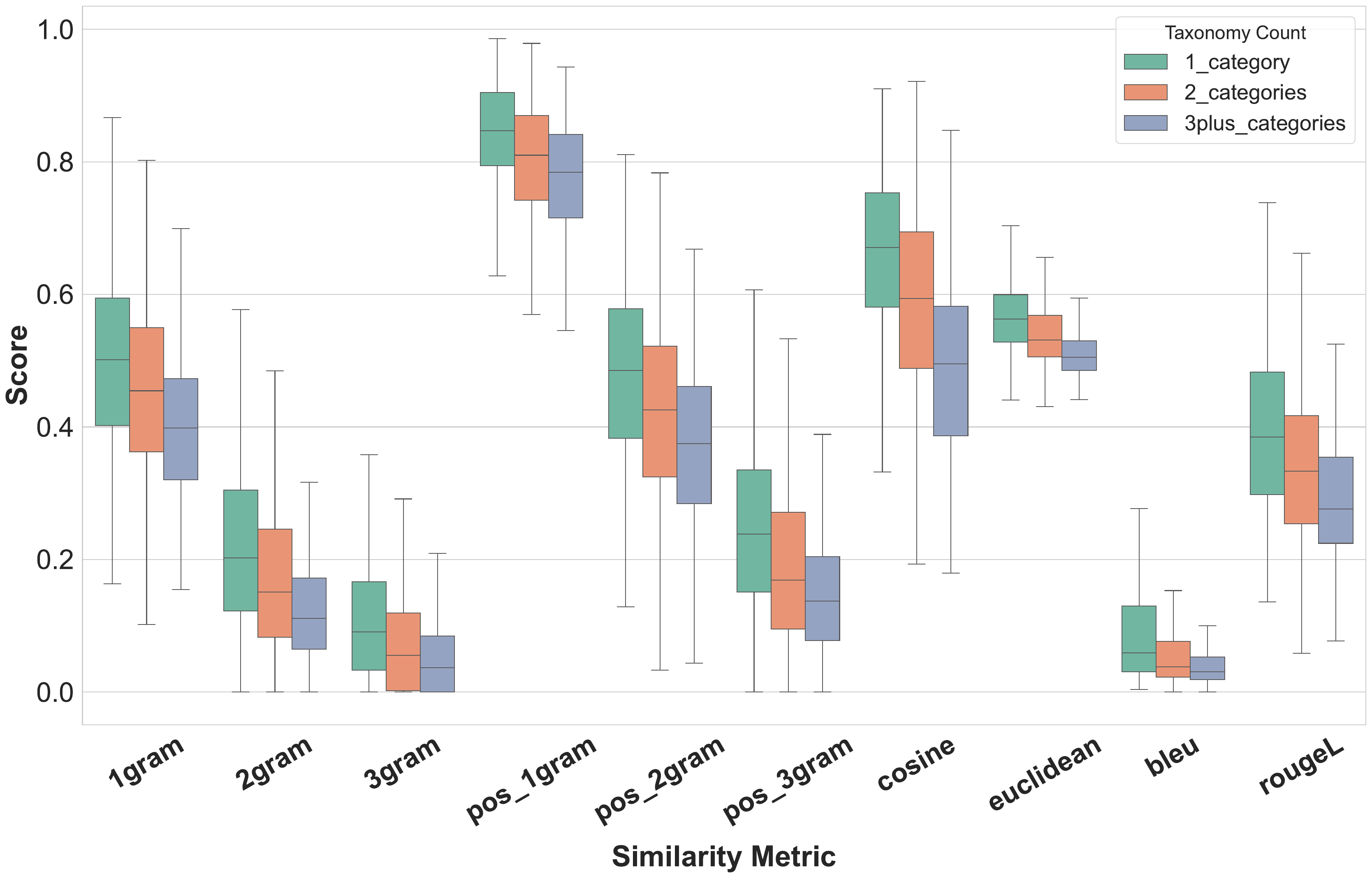}
    \caption{Boxplot of explanation similarities grouped by number of \taxonomy{} categories on an NLI item.
    }
    \label{fig:enter-label}
\end{figure}

To quantify it further, we compute pairwise similarity of explanations for each NLI item using standard metrics, following \citet{giulianelli-etal-2023-comes} and \citet{chen-etal-2025-rose}. These include 
lexical (word n-gram overlap),
morphosyntactic (POS n-gram overlap), and
semantic similarity (cosine/Euclidean distance\footnote{We compute semantic similarity using sentence embeddings from Sentence-BERT \citep{reimers-gurevych-2019-sentence}.}), along with BLEU \citep{papineni-etal-2002-bleu} and ROUGE-L \citep{lin-2004-rouge}. Figure~\ref{fig:enter-label} shows that explanation similarity decreases as the number of taxonomy categories increases, while explanations within the same category remain more similar, supporting the taxonomy’s ability to capture within-label variation.

\begin{figure}[t]
  \includegraphics[width=\columnwidth]{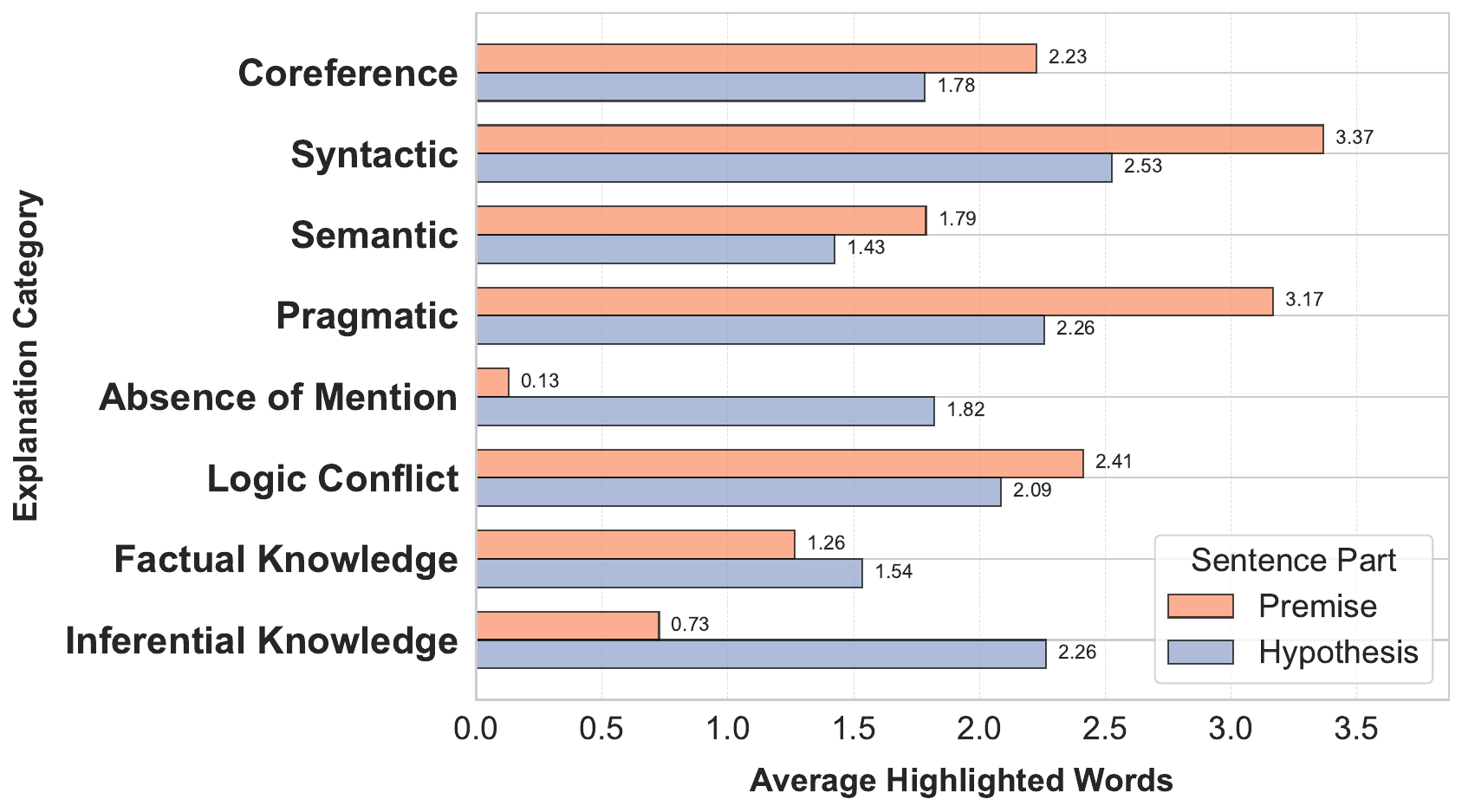}
  \caption{Average number of highlighted words in each premise-hypothesis pair across \taxonomy{} categories.}
  \label{fig:span_length}
\end{figure}

\paragraph{Highlights vs. Taxonomy}
\label{sec: highlightsvstax}
We analyze highlight span lengths for different explanation categories in Figure~\ref{fig:span_length}. On average, premises and hypotheses contain 13.81 and 7.41 words. \textit{Syntactic} explanations have the longest spans in both, reflecting sentence-level understanding. \textit{Absence of Mention} highlights are minimal in premises but more in hypotheses, marking new mentions in the hypotheses. \textit{Inferential} and \textit{Factual Knowledge} rely on short spans in the premise, pointing to external knowledge needs. These observations demonstrate that the length of highlight spans and distribution vary systematically across reasoning types, offering evidence that different types of reasoning reveal distinct linguistic patterns in NLI explanations.

\section{Explanation Generation using Taxonomy and Highlight}\label{sec:explanation-generation}

To investigate the interpretability and generalizability of our taxonomy, particularly in comparison to highlight approaches, we experiment on a practical usage: generating explanations with taxonomy or with highlight annotations. The goal is to generate, for a given NLI item and its label,  multiple explanations that reflect different plausible reasoning paths. While collecting such varied human-authored explanations is expensive—and often infeasible to elicit from a single annotator—LLMs offer a scalable alternative \cite{chen-etal-2025-rose}. 
We discuss various prompting paradigms (\S\ref{subsec:prompt}) and measure the similarities between LLM-generated and human explanations (\S\ref{subsec:generation-results}). 

\subsection{Prompting Paradigms}\label{subsec:prompt}

We experiment with three prompting paradigms and evaluate our approach on three instruction-tuned LLMs: GPT-4o, DeepSeek-v3, and Llama-3.3-70B-Instruct, with full prompt templates presented in Appendix~\ref{sec:appx-prompt-generation}.

\begin{table*}[t]
\centering
\resizebox{\textwidth}{!}{%
\begin{tabular}{@{}lccccccccccl@{}}
\toprule
\multirow{2}{*}{\textbf{Mode}} &
  \multicolumn{3}{c}{\textbf{Word n-gram}} &
  \multicolumn{3}{c}{\textbf{POS n-gram}} &
  \multicolumn{2}{c}{\textbf{Semantic}} &
  \multicolumn{2}{c}{\textbf{NLG Eval}} &
  \multirow{2}{*}{\textbf{Avg\_len}} \\
\cmidrule(lr){2-4} \cmidrule(lr){5-7}
\cmidrule(lr){8-9}
\cmidrule(lr){10-11}

&
  \textbf{1-gram} &
  \textbf{2-gram} &
  \textbf{3-gram} &
  \textbf{1-gram} &
  \textbf{2-gram} &
  \textbf{3-gram} &
  \textbf{Cos.} &
  \textbf{Euc.} &
  \textbf{BLEU} &
  \textbf{ROUGE-L} &
  \\ \midrule
GPT4o \textit{\, baseline}                & 0.291 & 0.117 & 0.049 & 0.882 & 0.488 & 0.226 & 0.556 & 0.524 & 0.051 & 0.272 & 24.995 \\
\textit{\, \, highlight (indexed)} & 0.402 & 0.124 & 0.053 & 0.878 & 0.481 & 0.222 & 0.554 & 0.522 & 0.051 & 0.269 & 28.240 \\
\textit{\, \, taxonomy (two-stage)}  & 0.418 & 0.128 & 0.071 & 0.886 & 0.495 & 0.242 & 0.593 & 0.537 & 0.071 & 0.314 & \textbf{19.991} \\
\textit{\, \, taxonomy (end-to-end)} & \textbf{0.437} & \textbf{0.166} & \textbf{0.083} & \textbf{0.898} & \textbf{0.511} & \textbf{0.255} & \textbf{0.608} & \textbf{0.540} & \textbf{0.074} & \textbf{0.323} & 26.672 \\
\midrule
DeepSeek-v3 \textit{\, baseline} & 0.369 & 0.087 & 0.034 & 0.847 & 0.449 & 0.195 & 0.428 & 0.490 & 0.042 & 0.245 & \textbf{20.288} \\
\textit{\, \,  highlight (indexed)} & 0.364 & 0.091 & 0.037 & 0.861 & 0.450 & 0.196 & 0.464 & 0.499 & 0.034 & 0.242 & 27.301 \\
\textit{\, \, taxonomy (two stage)} & 0.391 & 0.122 & 0.055 & 0.884 & 0.475 & 0.219 & 0.544 & 0.522 & 0.057 & 0.293 & 20.894 \\
\textit{\, \, taxonomy (end-to-end)} & \textbf{0.404} & \textbf{0.140} & \textbf{0.067} & \textbf{0.897} & \textbf{0.486} & \textbf{0.233} & \textbf{0.556} & \textbf{0.528} & \textbf{0.063} & \textbf{0.306} & 25.960 \\
\midrule
Llama-3.3-70B \textit{\, baseline} & 0.392 & 0.106 & 0.044 & 0.863 & 0.478 & 0.224 & 0.466 & 0.496 & 0.046 & 0.250 & 27.148 \\
\textit{\, \, highlight (indexed)} & 0.317 & 0.065 & 0.024 & 0.807 & 0.408 & 0.173 & 0.367 & 0.478 & 0.031 & 0.199 & 24.987 \\
\textit{\, \, taxonomy (two-stage)} & \textbf{0.444} & \textbf{0.167} & \textbf{0.082} & 0.889 & \textbf{0.512} & \textbf{0.256} & \textbf{0.609} & \textbf{0.541} & \textbf{0.078} & \textbf{0.321} & \textbf{22.340} \\
\textit{\, \, taxonomy (end-to-end)} & 0.383 & 0.110 & 0.048 & \textbf{0.896} & 0.499 & 0.232 & 0.505 & 0.510 & 0.047 & 0.262 & 28.870 \\
\bottomrule
\end{tabular}%
}
\caption{Similarity of LLM-generated explanations to human references. 
}
\label{tab:evaluation_0}
\end{table*}

\paragraph{Baseline} 
The model only sees the NLI item (premise and hypothesis) and a label, and generates explanations based on this input.

\paragraph{Highlight-Guided}
Adding to the baseline inputs, we include highlight annotations of the premise and hypothesis—as indices (\textit{indexed}) or tokens marked by surrounding \textbf{**} in text (\textit{in-text}). 
We ask the LLMs to first predict the highlighted tokens in the premise and hypothesis and subsequently generate relevant explanations.
We report results in the \textit{indexed} setup, as it yields marginally better average performance across metrics; see Appendix~\ref{sec:additional_results} for similar  \textit{in-text} results and when using e-SNLI highlights. 

\paragraph{Taxonomy-Guided}
The model is provided with the taxonomy description (Table~\ref{tab:taxonomy_questions}), one example for each of the eight reasoning categories, and the full taxonomy. We experiment with two prompting setups: \textit{two-stage} and \textit{end-to-end}. The \textit{two-stage} setup separates classification and generation—first predicting the taxonomy label for a given NLI item, then generating explanations conditioned on it. The \textit{end-to-end} approach performs both steps in a single prompt. This comparison addresses concerns that end-to-end generation may introduce a bias toward certain reasoning categories.

\subsection{Model Generation Results}\label{subsec:generation-results}

We evaluate similarities between LLM- and human-generated explanations using the same metrics as in \S\ref{subsec:variation-taxonomy-explanation}. For each generated explanation, we evaluate it against the human-written references individually by computing all metrics. We then select the best-scoring reference for that explanation and retain its score. The score for each NLI item is then obtained by averaging over all its generated explanations. The final reported result is the average of these per-item scores across our entire dataset.

Table~\ref{tab:evaluation_0} reports our generation results. Notably, \textit{end-to-end} taxonomy prompting performs best on GPT-4o and DeepSeek-v3, while \textit{two-stage} prompting yields better performance on Llama 3.3. Across all models, taxonomy-guided generation achieves higher alignment with human explanations than both the baseline and highlight-based approaches. This is reflected in higher POS tag n-gram overlap, which captures morphosyntactic structural similarity, and in stronger semantic similarity metrics like Cosine. In contrast, highlight-guided explanations perform comparably or slightly worse than the baseline, and tend to have longer average lengths with lower lexical and semantic overlap with the references. This suggests that highlighting alone may not sufficiently inform the model to produce relevant explanations. It is also worth noting that the open-source Llama model performs on par with the closed-source GPT model.

\begin{figure*}[t]
  \includegraphics[width=\textwidth]{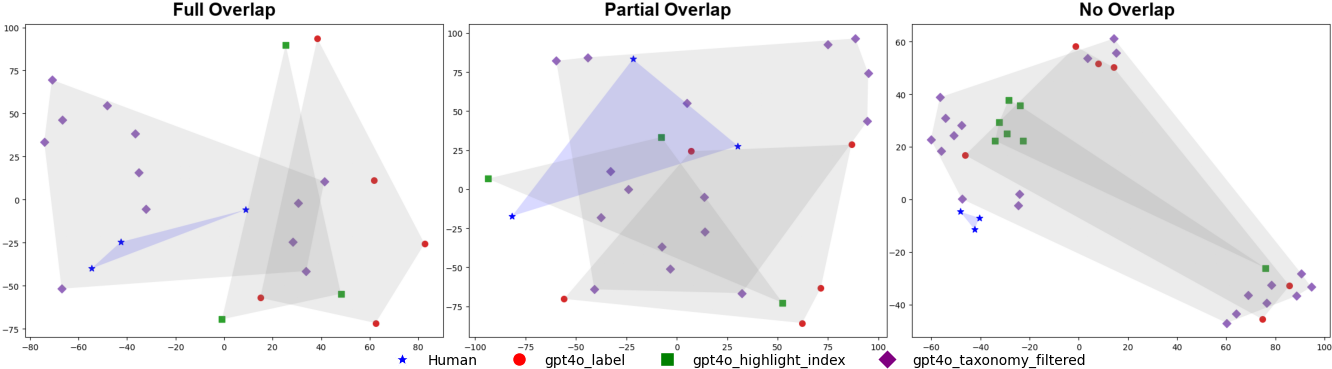}
  \caption{Representative t-SNE visualizations of explanation embeddings. The blue convex hull represents the span of human-written explanations, while the gray illustrates the spread of GPT4o-generated explanations.}
  \label{fig:visualization}
\end{figure*}

While high similarity to human references is desirable, overly verbose content may indicate unnecessary redundancy \cite{holtzman2020curiouscaseneuraltext}. From Table~\ref{tab:evaluation_0}, we observe that highlight-guided generations tend to produce longer explanations (e.g., 28.24 for GPT-4o and 30.42 for DeepSeek-v3) while yielding lower BLEU and ROUGE-L scores compared to both the baseline and taxonomy-guided variants. This indicates that the predicted highlights did not improve alignment with human-written explanations and may instead reflect redundancy.
Rather, taxonomy-based methods result in higher similarity and more concise explanations. This effect, however, is driven by the taxonomy two-stage approach: while it produces notably shorter outputs, the end-to-end taxonomy variant often generates longer explanations, in some cases exceeding those of highlight-based methods for Llama.

\subsection{Model Generation Validation}\label{subsec:generation-validation}
To assess the quality of model-generated NLI explanations, we conduct a round of human validation. Specifically, we evaluate 8,373 explanations produced by GPT-4o under the taxonomy \textit{two-stage} prompting paradigm introduced in §\ref{subsec:prompt}, as this setup yields a broader coverage of reasoning categories compared to the \textit{end-to-end} variant, allowing for more comprehensive validation across the taxonomy. For each explanation, one trained annotator is provided with the premise, hypothesis, NLI label, the generated explanation, and the corresponding taxonomy category. The annotator is instructed to answer the following two binary questions:

\begin{enumerate}
    \item NLI label consistency: Does the explanation fit the gold label? (Yes/No)
    \item Taxonomy consistency: Does the explanation fit the taxonomy? (Yes/No)
\end{enumerate}

The results show that overall 98.27\% of the generated explanations align with the NLI label and 83.84\% match the prompted taxonomy category. Some categories showed high alignment rates, such as \textit{Syntactic} (94.88\%) and \textit{Absence of Mention} (92.47\%), while others were more challenging, such as \textit{Coreference} (57.25\%) and \textit{Logic Conflict} (63.05\%). A detailed breakdown by taxonomy category, along with further discussion, is provided in Appendix~\ref{sec:human-validation}.

\section{Assessing Explanation Coverage: Human vs. LLM Outputs} 
\label{sec:coverage}

Besides evaluating the similarity between human-written and LLM-generated explanations, the more fundamental question is \textit{how much within-label variation can LLM-generated explanations capture.}
Are LLMs too repetitive and only cover a subset of human explanations?
Can LLMs unearth appropriate new explanations that are missing from a few human-written ones?
This section presents our attempt to measure coverage in LLM-generated explanations.
Given that LLMs are prompted to generate multiple explanations, we examine whether they can fully cover the semantic space of human explanations and potentially extend beyond it.

Figure~\ref{fig:visualization} illustrates this semantic coverage for three representative instances from \dataset{}. From left to right, the examples demonstrate: (1) full coverage, where the convex hull of model-generated explanations fully encloses the human explanation points; (2) partial coverage, where model generations cover some of the human reference points and (3) no coverage, where model outputs cover no human explanation point.

\begin{table}[t]
    \centering
    \footnotesize
    \resizebox{\columnwidth}{!}{%
    \begin{tabular}{lrrrr}
        \toprule
        & \multicolumn{2}{c}{\textbf{Coverage }} & \multicolumn{2}{c}{\textbf{Area}} \\
        \textbf{Mode} & \textbf{Full} & \textbf{Partial} & \textbf{Rec} & \textbf{\makecell{Prec}}\\ \midrule
        GPT4o
  \textit{baseline} & 1.9 & 21.6 & 16.5 & \textbf{5.7} \\ 
     \, \,    \textit{highlight (indexed)} & 1.1  & 13.5  & 10.0 & 4.7 \\ 
      \, \,   \textit{taxonomy (end-to-end)} & \textbf{10.7}  & \textbf{56.1}  & \textbf{49.3}  & 5.6 \\ \midrule
        DeepSeek-v3 
        \textit{baseline} & 4.0  & 20.5  & 17.5 & 2.7 \\ 
     \, \,    \textit{highlight (indexed)} & 2.3  & 14.9  & 12.5 & 2.9 \\ 
     \, \,    \textit{taxonomy (end-to-end)} & \textbf{17.8}  & \textbf{61.8}  & \textbf{54.7} & \textbf{3.8} \\ \midrule
        Llama-3.3-70B 
        \textit{baseline} & 1.7  & 15.4 & 12.2 & 2.9 \\ 
    \, \,     \textit{highlight (indexed)} & 0.5  & 8.2  & 6.5 & 2.5 \\ 
    \, \,     \textit{taxonomy (end-to-end)} & \textbf{16.7}  & \textbf{65.2}  & \textbf{59.8} & \textbf{5.7} \\ \bottomrule
    \end{tabular}
    }
    \caption{Results on the semantic coverage of model explanations regarding human reference explanations.
    }
    \label{tab:visualization_analysis}
\end{table}

\begin{table*}[t!hb]
  \centering
  \footnotesize
  \resizebox{\textwidth}{!}{%
  \begin{tabularx}{\textwidth}{lX}
  \toprule
  \textbf{Premise:} & \textit{A person wearing a straw hat, standing outside working a steel apparatus with a pile of coconuts on the ground.}\\
  \textbf{Hypothesis:} & \textit{A person is near a pile of coconuts.}\\
  \textbf{Label:} & Entailment \\
  \midrule
  \textbf{Source} & \textbf{Explanations} \\
  \midrule
  \textbf{Human} &
  - With a pile of coconuts implies near a pile of coconuts. \textcolor{purple}{[Semantic]} \newline
  - A person standing near a pile of coconuts would be standing outside. \textcolor{purple}{[Inferential Knowledge]} \newline
  - If the person is with the pile of coconuts on the ground, they are near them. \textcolor{purple}{[Inferential Knowledge]} \\
  \midrule
  \textbf{Baseline} &
  - There is no indication in the content that the person and the coconuts are separated by any significant distance. \newline
  - The use of ``near'' in the statement aligns with the spatial arrangement described in the content, where the person and coconuts are part of the same setting. \newline
  - The person is described as standing outside, suggesting they are near the objects in the scene. \\
  \midrule
  \textbf{Highlight} &
  - The phrase ``standing outside'' in the content implies that the person is near the ground where the pile of coconuts is located, supporting the statement that the person is near the pile of coconuts. \textcolor{blue}{["12,13,14,15,16", "3,4,5,6,7"]} \newline
  - The content describes ``a pile of coconuts on the ground,'' which directly correlates to the statement’s ``a pile of coconuts,'' indicating the same pile is being referenced. \textcolor{blue}{["6,7", "3"]} \newline
  - The person ``working a steel apparatus'' in the content might suggest proximity to the pile of coconuts since the work being done is likely related to the coconuts, thus supporting the statement that the person is near them. \textcolor{blue}{["1,12,16,19", "1,3,7"]} \\
  \midrule
  \textbf{Taxonomy} &
  - Standing outside implies proximity to the objects on the ground, including the pile of coconuts. \textcolor{blue}{[Inferential Knowledge]} \newline
  - The word `near' denotes closeness, which is implied by the person standing in the vicinity of the coconuts. \textcolor{blue}{[Semantic]} \newline
  - The hypothesis simplifies the structure of the premise by focusing on the person’s proximity to the coconuts, which is implied by the premise. \textcolor{blue}{[Syntactic]} \\
  \bottomrule
  \end{tabularx}
  }
  \caption{Explanations from different generation strategies for one \dataset\  item. 
  For human explanations, annotator-assigned categories are in \textcolor{purple}{purple}.
  Model-generated taxonomy categories and highlight indexes are in \textcolor{blue}{blue}.}
  \label{tab:explanation_examples}
\end{table*}

\paragraph{Proposed Measures}
We propose four measures, \textit{full coverage}, \textit{partial coverage}, \textit{area precision}, and \textit{area recall} to analyze the semantic space between model- and human-generated explanations using t-SNE visualizations and convex hull statistics \cite{JMLR:v9:vandermaaten08a}. 

We define \textit{full coverage} as a binary condition: an NLI item is fully covered (yes) if all human explanation reference points fall within the convex hull spanned by the model explanations, and not covered (no) otherwise.
Similarly, it is \textit{partially covered} if at least one human reference point is within the model explanation space. 
\textit{Full and partial coverage} computes the percentage of 1,002 \dataset{} items whose explanations are fully or partially covered within the convex hull of the model explanations. 

On the other hand, \textit{area precision and recall} assess for each NLI item, the overlapping area between the space spanned by all reference explanations and that spanned by all model explanations. 
\textit{Area precision} measures the ratio of the overlapping area over the area spanned by model explanations, and \textit{area recall} over the area spanned by human explanations. 
We report the average of \textit{area precision} and \textit{area recall} over 1,002 instances. 

\paragraph{Results}
Table~\ref{tab:visualization_analysis} shows that taxonomy-guided explanation generation consistently achieves the highest full and partial coverage of reference explanation points.
They also yield the highest average area recall and precision, in all test cases except the GPT4o baseline, indicating that the semantic space overlap between taxonomy-guided model explanations and human explanations is large.

In contrast, baseline and highlight-guided modes show much lower full and partial coverage and smaller overlap ratios. It indicates that the explanation spaces are less aligned with human explanations. Although highlight-guided outputs tend to form smaller and more concentrated explanation regions (as seen in their low area precision), this compactness does not mean their explanations are more meaningful. When guided by highlights, the model often fails to generate explanations that reflect the essential ideas expressed by humans. These results highlight that prompting using taxonomy-based guidance is more effective at generating human-aligned explanations in the embedding space.

We observe in Table~\ref{tab:visualization_analysis} that GPT4o exhibits lower coverage compared to DeepSeek and Llama, which can partly be attributed to the smaller number of explanations generated per instance. In our setup, the models are prompted to produce all possible explanations given an NLI instance, a label, and optionally a taxonomy category or relevant highlights. On average, GPT4o generates 3.59 explanations per example, while DeepSeek and Llama produce 5.90 and 6.14, respectively. This difference in output quantity naturally contributes to GPT4o’s lower coverage. 

\paragraph{Case Study}
Table~\ref{tab:explanation_examples} provides a concrete example (the leftmost case in Figure~\ref{fig:visualization}) where the human explanations are fully covered by the taxonomy-guided generation but only partially captured by label- and highlight-guided generations.

Human explanations focus on spatial proximity (\textit{near}) and real-world expectations (i.e.,  coconuts being outdoors). 
The baseline and highlight-guided explanations also refer to the spatial proximity. However, the reasoning is less precise and often vague, lacking the structure seen in human explanations. 
Instead, taxonomy-guided generations are not only more coherent and concise, but also cover a broader range of reasoning types.
In addition to producing outputs aligned with \textit{Semantic} and \textit{Inferential Knowledge}, they provide an additional \textit{Syntactic}-labeled explanation, addressing the sentence simplification from premise to hypothesis.
 
However, while the taxonomy-generated explanation ``standing outside implies proximity to the objects on the ground, including the pile of coconuts'' captures the essence of the human-written ``a person standing near a pile of coconuts would be standing outside,'' it is more abstract and less natural when expressing the casual contexts.
All generated explanations, particularly highlight-guided ones, are also longer than the human-written ones, echoing the redundancy issue discussed in \S\ref{subsec:generation-results}.

\section{Conclusion}
In this work, we introduce \taxonomy{}, a linguistically-informed taxonomy designed to capture different reasoning strategies behind NLI explanations, with a particular focus on within-label variation. 
The learnability evaluation 
shows that models, after fine-tuning or few-shot prompting, can effectively classify explanations into our taxonomy, demonstrating its practicality. 
We further demonstrate that taxonomy guidance consistently helps generation, resulting in model explanations that are semantically richer and closer to human explanations than baseline or highlight-based approaches.

Overall, our work bridges human reasoning strategies and model predictions in a structured way, providing a foundation for more interpretable NLI modeling. In addition, we enhance the e-SNLI dataset with fine-grained taxonomy categories for explanations, providing a resource to support future work.
While our current evaluation focuses on a specific subset of NLI data, future work will extend this approach to 
broader variation-aware benchmarks such as ANLI~\cite{nie-etal-2020-adversarial} and LiveNLI~\cite{jiang_ecologically_2023}. These extensions will enable a more comprehensive assessment of the taxonomy’s generalizability across diverse inference settings. Annotations, generated explanations, and code will be released publicly upon publication.\!\footnote{Dataset and implementation are publicly available at \href{https://github.com/mainlp/LiTEx}{https://github.com/mainlp/LiTEx} for reproduction.}

\section*{Limitations}

While our taxonomy offers a structured and linguistically informed perspective to analyze different types of explanation in NLI, it has several limitations. First, the annotation process, though guided by detailed definitions, still involves subjective interpretation from a single annotator, especially for borderline categories such as \textit{Factual Knowledge} versus \textit{Inferential Knowledge}. This highlights the inherent subjectivity of explanation annotation, where different annotators may reasonably disagree on the most appropriate category. Second, our taxonomy focuses solely on explicit explanations provided in natural language. It does not account for the implicit reasoning process that may not be verbalized in text. This may limit the taxonomy’s applicability to inferred or implied reasoning, especially when applying it to other NLI datasets without free-text explanations. Finally, our current experiments are conducted on the e-SNLI dataset, which may not represent the full spectrum of natural language inference.

\section*{Ethical considerations} 
We do not foresee any ethical concerns associated with this work. All analyses were conducted using publicly available datasets and models. No private or sensitive information was used. Additionally, we will release our code, prompts, and documentation to support transparency and reproducibility.

\section*{Acknowledgments}

We thank the members of the MaiNLP lab for their insightful feedback on earlier drafts of this paper. 
We specifically appreciate the suggestions of Verena Blaschke, Andreas Säuberli, and Yang Janet Liu. 
Beiduo Chen acknowledges his membership in the European Laboratory for Learning and Intelligent Systems (ELLIS) PhD program.
Marie-Catherine de Marneffe is a Research Associate of
the Fonds de la Recherche Scientifique – FNRS.
This research is supported by ERC Consolidator Grant DIALECT 101043235.

\paragraph{Use of AI Assistants} The authors acknowledge the use of ChatGPT solely for assistance with grammar, punctuation, and vocabulary corrections, as well as for supporting coding tasks.


\appendix

\section{Illustrative Examples of the Taxonomy}
\label{sec:Taxonomy_examples}

This section provides illustrative examples to clarify and exemplify our taxonomy. For each example, we present the premise, hypothesis, and human explanation as they appear in the original dataset, preserving all original text, including any typos or grammatical errors. In Table~\ref{tab:taxonomy_example1} and Table~\ref{tab:taxonomy_example2}, two representative examples are listed for the two broad categories: \textit{Text-Based (TB) Reasoning} and \textit{World-Knowledge (WK) Reasoning}.

\begin{table*}[h]
  \centering
  \small
  \resizebox{\textwidth}{!}{%
  \begin{tabularx}{\textwidth}{lX}
    \toprule
    \multicolumn{2}{l}{\textbf{Coreference}} \\
    \midrule
    Premise: & The man in the black t-shirt is trying to throw something. \\
    Hypothesis: & The man is in a black shirt. \\
    Gold Label: & Entailment \\
    Explanation: & The man is in a black shirt refers to the man in the black t-shirt. \\
    \midrule 
    Premise: & A naked man rides a bike. \\
    Hypothesis: & A person biking. \\
    Gold Label: & Entailment \\
    Explanation: & The person biking in the hypothesis is the naked man. \\
    \midrule \midrule 
    \multicolumn{2}{l}{\textbf{Semantic}} \\
    \midrule
    Premise: & A man in a black tank top is wearing a red plaid hat. \\
    Hypothesis: & A man in a hat. \\
    Gold Label: & Entailment \\
    Explanation: & A red plaid hat is a specific type of hat. \\
    \midrule 
    Premise: & Three man are carrying a red bag into a boat with another person and boat in the background. \\
    Hypothesis: & Some people put something in a boat in a place with more than one boat. \\
    Gold Label: & Entailment \\
    Explanation: & Three men are people. \\
    \midrule \midrule 
    \multicolumn{2}{l}{\textbf{Syntactic}}\\
    \midrule
    Premise: & Two women walk down a sidewalk along a busy street in a downtown area. \\
    Hypothesis: & The women were walking downtown. \\
    Gold Label: & Entailment \\
    Explanation: & The women were walking downtown is a rephrase of, Two women walk down a sidewalk along a busy street in a downtown area. \\
    \midrule 
    Premise: & Bruce Springsteen, with one arm outstretched, is singing in the spotlight in a dark concert hall. \\
    Hypothesis: & Bruce Springsteen is a singer. \\
    Gold Label: & Entailment \\
    Explanation: & Springsteen is singing in a concert hall. \\
    \midrule \midrule
    \multicolumn{2}{l}{\textbf{Pragmatic}}\\
    \midrule
    Premise: & A girl in a blue dress takes off her shoes and eats blue cotton candy. \\
    Hypothesis: & The girl is eating while barefoot. \\
    Gold Label: & Entailment \\
    Explanation: & If a girl takes off her shoes, then she becomes barefoot, and if she eats blue candy, then she is eating. \\
    \midrule 
    Premise: & A woman wearing bike shorts and a skirt is riding a bike and carrying a shoulder bag. \\
    Hypothesis: & A woman on a bike. \\
    Gold Label: & Entailment \\
    Explanation: & Woman riding a bike means she is on a bike \\
    \midrule \midrule
    \multicolumn{2}{l}{\textbf{Absence of Mention}}\\
    \midrule
    Premise: & A person with a purple shirt is painting an image of a woman on a white wall. \\
    Hypothesis: & A woman paints a portrait of a person. \\
    Gold Label: & Neutral \\
    Explanation: & A person with a purple shirt could be either a man or a woman. We can't assume the gender of the painter. \\
    \midrule 
    Premise: & A young man in a heavy brown winter coat stands in front of a blue railing with his arms spread. \\
    Hypothesis: & The railing is in front of a frozen lake. \\
    Gold Label: & Neutral \\
    Explanation: & It does not say anything about there being a lake. \\
    \midrule \midrule
    \multicolumn{2}{l}{\textbf{Logic Conflict}} \\
    \midrule
    Premise: & Five girls and two guys are crossing an overpass. \\
    Hypothesis: & The three men sit and talk about their lives. \\
    Gold Label: & Contradiction \\
    Explanation: & Three is not two. \\
    \midrule 
    Premise: & Many people standing outside of a place talking to each other in front of a building that has a sign that says ``HI-POINTE''. \\
    Hypothesis: & The group of people aren't inside of the building.\\
    Gold Label: & Entailment \\
    Explanation: & The people described are standing outside, so naturally not inside the building. \\
    \bottomrule
  \end{tabularx}
  }
  \caption{Illustrative examples of the taxonomy (Text-Based Reasoning).}
  \label{tab:taxonomy_example1}
\end{table*}

\begin{table*}[ht]
  \centering
  \small
  \resizebox{\textwidth}{!}{%
  \begin{tabularx}{\textwidth}{lX}
    \toprule
    \multicolumn{2}{l}{\textbf{Factual Knowledge}} \\
    \midrule
    Premise: & Two people crossing by each other while kite surfing. \\
    Hypothesis: & The people are both males. \\
    Gold Label: & Neutral \\
    Explanation: & Not all people are males. \\
    \midrule 
    Premise: & Here is a picture of people getting drunk at a house party. \\
    Hypothesis: & Some people are by the side of a swimming pool party. \\
    Gold Label: & Neutral \\
    Explanation: & Not all houses have swimming pools. \\
    \midrule \midrule 
    \multicolumn{2}{l}{\textbf{Inferential Knowledge}} \\
    \midrule
    Premise: & A girl in a blue dress takes off her shoes and eats blue cotton candy. \\
    Hypothesis: & The girl in a blue dress is a flower girl at a wedding. \\
    Gold Label: & Neutral \\
    Explanation: & A girl in a blue dress doesn’t imply the girl is a flower girl at a wedding. \\
    \midrule 
    Premise: & A person dressed in a dress with flowers and a stuffed bee attached to it, is pushing a baby stroller down the street. \\
    Hypothesis: & An old lady pushing a stroller down a busy street. \\
    Gold Label: & Neutral \\
    Explanation: & A person in a dress of a particular type need neither be old nor female.  A street need not be considered busy if only one person is pushing a stroller down it. \\
    \bottomrule
  \end{tabularx}
  }
  \caption{Illustrative examples of the taxonomy (World Knowledge-Based Reasoning).}
  \label{tab:taxonomy_example2}
\end{table*}

\FloatBarrier

\section{Taxonomy Validation: IAA Classification Report}
\label{sec:IAA-report}

\begin{figure}[h]
  \includegraphics[width=\columnwidth]{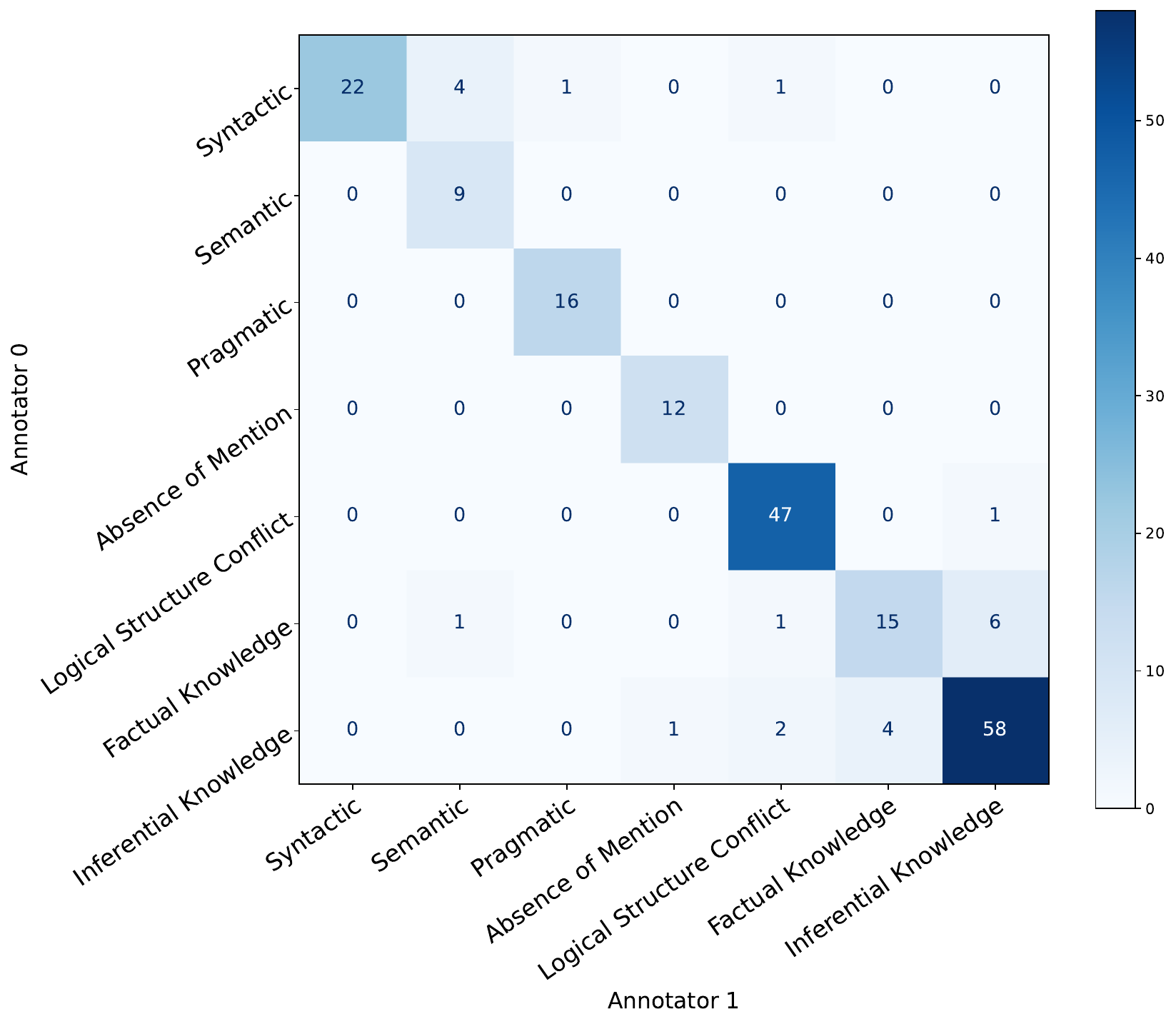}
  \caption{Inter-Annotator Confusion Matrix for Explanation Category Annotation.}
  \label{fig:confusion}
\end{figure}

Figure~\ref{fig:confusion} presents the inter-annotator confusion matrix for explanation category annotation, used to validate the proposed taxonomy. Overall, we observe strong agreement across most categories, with especially high consistency in categories such as \textit{Logic Conflict} and \textit{Inferential Knowledge}. Some confusion appears between semantically adjacent categories, such as \textit{Factual Knowledge} vs. \textit{Inferential Knowledge}, and \textit{Semantic} vs. \textit{Syntactic}.

To better understand the nature of inter-annotator disagreement in our taxonomy-based labeling, we present a qualitative analysis of several items with mismatched labels in the confusion matrix. The following examples shed light on how subtle differences in reasoning can lead to divergent category assignments:\\ \\
\colorbox{gray!10}{\textit{Factual Knowledge vs. Logic Conflict}} \\
\textbf{Premise:} An old man with a package poses in front of an advertisement. \\
\textbf{Hypothesis:} A man walks by an ad. \\
\textbf{Explanation:} Poses is different from walks. \\
\textbf{Category (Annotator 0):} Factual Knowledge \\
\textbf{Category (Annotator 1):} Logic Conflict \\
\textbf{Analysis:} Annotator 0 likely interprets the explanation as highlighting a factual discrepancy in the physical action ``posing'' vs.``walking''), treating this as a knowledge-based distinction about what the person is doing. Annotator 1, on the other hand, may view the same contrast as introducing a logical inconsistency in the event semantics—i.e., the man cannot be simultaneously posing and walking, which reflects a conflict in entailment assumptions. This illustrates how borderline cases between fact-based knowledge and event logic can be interpreted differently, especially when both literal and inferential mismatches are present.\\ \\
\colorbox{gray!10}{\textit{Inferential vs. Factual Knowledge}} \\
\textbf{Premise:} A young family enjoys feeling ocean waves lap at their feet. \\
\textbf{Hypothesis:} A young man and woman take their child to the beach for the first time. \\
\textbf{Explanation:} The young family does not mean that they have a child at the beach. \\
\textbf{Category (Annotator 0):} Inferential Knowledge \\
\textbf{Category (Annotator 1):} Factual Knowledge \\
\textbf{Analysis:} Annotator 0 interprets the inference from ``young family'' to ``having a child present'' as requiring reasoning with world knowledge about family structures. In contrast, Annotator 1 views this as an incorrect factual claim, where the hypothesis wrongly assumes a child is present. This disagreement highlights the challenge of distinguishing between inferential reasoning and factual correction, indicating a need for clearer taxonomy boundaries. \\ \\
\colorbox{gray!10}{\textit{Syntactic vs. Semantic}} \\
\textbf{Premise:} Two children are laying on a rug with some wooden bricks laid out in a square between them. \\
\textbf{Hypothesis:} Two children are on a rug. \\
\textbf{Explanation:} To say the children are ``laying on'' a rug is rephrasing ``on'' a rug. \\
\textbf{Category (Annotator 0):} Syntactic \\
\textbf{Category (Annotator 1):} Semantic \\
\textbf{Analysis:} Annotator 0 classifies the change from ``laying on'' to ``on'' as a simple syntactic variation, treating it as a surface-level rewording. In contrast, Annotator 1 interprets this shift as semantically meaningful, possibly inferring that ``laying on'' conveys posture or state, thus labeling it as a Semantic shift. This disagreement illustrates a key challenge in NLI: distinguishing between purely syntactic paraphrases and cases where subtle wording changes alter meaning. Such distinctions become especially nuanced when modifications involve minor phrasing differences.

\section{Taxonomy Validation: LM and LLM Classification}
\label{sec:classification-details}

In Table~\ref{tab:hyperparameter} the hyperparameter setup of fine-tuning BERT and RoBERTa is listed. 
We follow a standard supervised classification pipeline, where the model takes as input the concatenated premise, hypothesis, label, and explanation, and predicts the correct explanation category among eight categories. For validation, we measured both the classification accuracy and the macro-F1 score across the explanation categories, as shown in Table~\ref{tab:fine-tune-results}. We selected the best-performing checkpoint based on the highest macro-F1 on the dev set for final evaluation.

\begin{table*}[h]
\centering
\small
\begin{tabular}{llcccccc}
\toprule
\multicolumn{1}{c}{} &
  \multicolumn{1}{l}{} &
  \multicolumn{3}{c}{\textit{\textbf{roberta-base}}} &
  \multicolumn{3}{c}{\textbf{bert-base}} \\ 
\multicolumn{1}{c}{\multirow{-2}{*}{\textbf{Explanation Category}}} &
  \multicolumn{1}{l}{\multirow{-2}{*}{\textbf{\begin{tabular}[c]{@{}l@{}}data split\\ (\textit{train/dev/test})\end{tabular}}}} &
  \textbf{Precision} & \textbf{Recall} & \multicolumn{1}{l}{\textbf{F1}} & \textbf{Precision} & \textbf{Recall} & \textbf{F1} \\ \midrule
 &
  \multicolumn{1}{l|}{40/20/40} & 0.00 & 0.00 & \multicolumn{1}{c|}{0.00} &
    0.00 & 0.00 & 0.00 \\
  \multirow{-2}{*}{Coreference} &
  \multicolumn{1}{l|}{50/0/50} & 1.00 & 0.04 & \multicolumn{1}{c|}{0.07} &
  0.00 & 0.00 & 0.00 \\ \midrule
 &
  \multicolumn{1}{l|}{40/20/40} & 0.58 & 0.63 & \multicolumn{1}{c|}{0.61} & 
  0.54 & 0.68 & 0.61 \\
  \multirow{-2}{*}{Semantic} & \multicolumn{1}{l|}{50/0/50} & 0.57 & 0.68 & \multicolumn{1}{c|}{0.62} & 
  0.54 & 0.64 & 0.59 \\ \midrule
 &
  \multicolumn{1}{l|}{40/20/40} & 0.64 & 0.74 & \multicolumn{1}{c|}{0.68} &
  0.61 & 0.77 & 0.68 \\
\multirow{-2}{*}{Syntactic} &
  \multicolumn{1}{l|}{50/0/50} & 0.62 & 0.76 & \multicolumn{1}{c|}{0.69} &
  0.62 & 0.80 & 0.69 \\ \midrule
 &
  \multicolumn{1}{l|}{40/20/40} & 0.53 & 0.74 & \multicolumn{1}{c|}{0.62} &
  0.57 & 0.65 & 0.61 \\
\multirow{-2}{*}{Pragmatic} &
  \multicolumn{1}{l|}{50/0/50} & 0.59 & 0.63 & \multicolumn{1}{c|}{0.61} & 
  0.60 & 0.58 & 0.59 \\ \midrule
 &
  \multicolumn{1}{l|}{40/20/40} & 1.00 & 0.23 & \multicolumn{1}{c|}{0.38} &
  0.95 & 0.42 & 0.58 \\
\multirow{-2}{*}{Absence of Mention} & 
  \multicolumn{1}{l|}{50/0/50} & 0.93 & 0.52 & \multicolumn{1}{c|}{0.67} &  
  0.96 & 0.41 & 0.57 \\ \midrule
 &
  \multicolumn{1}{l|}{40/20/40} & 0.81 & 0.83 & \multicolumn{1}{c|}{0.82} &
  0.78 & 0.87 & 0.82 \\
\multirow{-2}{*}{Logic Conflict} & 
  \multicolumn{1}{l|}{50/0/50} & 0.81 & 0.83 & \multicolumn{1}{c|}{0.82} &
  0.78 & 0.88 & 0.83 \\ \midrule
 &
  \multicolumn{1}{l|}{40/20/40} & 0.61 & 0.51 & \multicolumn{1}{c|}{0.55} &
  0.57 & 0.50 & 0.53 \\
\multirow{-2}{*}{Factual Knowledge} &
  \multicolumn{1}{l|}{50/0/50} & 0.62 & 0.55 & \multicolumn{1}{c|}{0.59} &
  0.61 & 0.56 & 0.58 \\ \midrule
 &
  \multicolumn{1}{l|}{40/20/40} & 0.75 & 0.81 & \multicolumn{1}{c|}{0.78} &
  0.79 & 0.76 & 0.77 \\
\multirow{-2}{*}{Inferential Knowledge} &
  \multicolumn{1}{l|}{50/0/50} & 0.79 & 0.82 & \multicolumn{1}{c|}{0.80} &
  0.80 & 0.79 & 0.80 \\
\multicolumn{8}{c}{\cellcolor[HTML]{EFEFEF}\textit{\textbf{Summary}}} \\
\multicolumn{1}{c}{} & \multicolumn{1}{l|}{40/20/40} & \multicolumn{3}{c|}{0.67} & \multicolumn{3}{c}{0.70} \\
  \multicolumn{1}{c}{\multirow{-2}{*}{acuuracy}} & \multicolumn{1}{l|}{50/0/50} &\multicolumn{3}{c|}{0.67} &\multicolumn{3}{c}{0.70} \\ \midrule
\multicolumn{1}{c}{} & \multicolumn{1}{l|}{40/20/40} & 0.47 & 0.49 & \multicolumn{1}{c|}{0.47} &
  0.60 & 0.58 & 0.58 \\ 
\multicolumn{1}{c}{\multirow{-2}{*}{maro avg}} & \multicolumn{1}{l|}{50/0/50} & 0.48 & 0.53 & \multicolumn{1}{c|}{0.50} &
  0.61 & 0.58 & 0.58 \\ \midrule
\multicolumn{1}{c}{} & \multicolumn{1}{l|}{40/20/40} & 0.61 & 0.67 & \multicolumn{1}{c|}{0.64} &
  0.68 & 0.70 & 0.68 \\ 
\multicolumn{1}{c}{\multirow{-2}{*}{weighted}} &
  \multicolumn{1}{l|}{50/0/50} & 0.65 & 0.69 & \multicolumn{1}{c|}{0.66} &
  0.68 & 0.70 & 0.69 \\ \midrule
\end{tabular}
\caption{RoBERTA and BERT fine-tuning results.}
\label{tab:fine-tune-results}
\end{table*}

We design a set of experiments to assess the ability of LLMs to classify NLI explanations into one of eight fine-grained categories (introduced in Section~\ref{sec:taxonomy}). Our evaluation covers zero-shot prompting (no training examples), one-shot prompting (a single annotated example), and few-shot prompting (two examples per category). A consistent prompting strategy is applied across models, with all templates provided in Table~\ref{tab:classification-prompt}.

\begin{table}[H]
\centering
\small
\begin{tabular}{lll}
\toprule
\textbf{Hyperparameter} & \textbf{BERT} & \textbf{RoBERTa} \\ \midrule
Learning Rate Decay     & Linear        & Linear           \\
Weight Decay            & 0.0           & 0.0              \\
Optimizer               & AdamW         & AdamW            \\
Adam $\epsilon$         & 1e-8          & 1e-8             \\
Adam $\beta_1$          & 0.9           & 0.9              \\
Adam $\beta_2$          & 0.999         & 0.999            \\
Warmup Ratio            & 0\%           & 0\%              \\
Learning Rate           & 2e-5          & 3e-5             \\
Batch Size              & 8             & 8                \\
Num Epoch               & 4             & 3                \\ \bottomrule
\end{tabular}
\caption{Hyperparameter used for fine-tuning BERT and RoBERTa models.}
\label{tab:hyperparameter}
\end{table}

\begin{table*}[t]
\centering
\footnotesize
\begin{tabularx}{\textwidth}{@{}l>{\raggedright\arraybackslash}X@{}}
\toprule
\textbf{Mode} & \textbf{General Instruction Prompt} \\ \midrule
\textit{without instruction and example} &
\textbf{``role'': ``user'', ``content'':}

You are an expert in solving Natural Language Inference tasks. Your task is to classify the following explanations into one of the categories listed below. Each category reflects a specific type of inference in the explanation between the premise and hypothesis. Here are the categories:

1. Coreference

2. Syntactic

3. Semantic

4. Pragmatic

5. Absence of Mention

6. Logic Conflict

7. Factual Knowledge

8. Inferential Knowledge
\\[2ex] \midrule
\textit{+ instruction} &
\textbf{``role'': ``user'', ``content'':}

You are an expert in solving Natural Language Inference tasks. Your task is to classify the following explanations into one of the categories listed below. Each category reflects a specific type of inference in the explanation between the premise and hypothesis.

Here are the categories:

1. Coreference
    - The explanation resolves references (e.g., pronouns or demonstratives) across premise and hypothesis. 

2. Syntactic
    - Based on structural rephrasing with the same meaning (e.g., syntactic alternation, coordination, subordination). If the explanation itself is the rephrasing of the premise or hypothesis, it should be included in this category. 
    
3. Semantic
    - Based on word meaning (e.g., synonyms, antonyms, negation).

4. Pragmatic
    - This category would capture inferences that arise from logical implications embedded in the structure or semantics of the text itself, without relying on external context or background knowledge.

5. Absence of Mention
    - Lack of supporting evidence, the hypothesis introduces information that is not supported, not entailed, or not mentioned in the premise, but could be true.

6. Logic Conflict
    - Structural logical exclusivity (e.g., either-or, at most, only, must), quantifier conflict, temporal conflict, location conflict, gender conflict etc. 

7. Factual Knowledge
    - Explanation relies on common sense, background, or domain-specific facts. No further reasoning involved. 

8. Inferential Knowledge
    - Requires real-world causal, probabilistic reasoning or unstated but assumed information. 

Respond **only with the number (1–8)** corresponding to the most appropriate category. 
\\[2ex] 

\bottomrule
\end{tabularx}
\caption{Instruction prompts for LLMs as classifiers.}
\label{tab:classification-prompt}
\end{table*}

Specifically, we experiment with Llama-3.2-3B-Instruct \cite{llama_3}, GPT-3.5-turbo \cite{brown_language_2020}, GPT-4o \cite{openai_gpt-4_2024}, and DeepSeek-v3 \cite{deepseekai2025deepseekv3technicalreport}, under six experimental configurations:

\begin{enumerate}[nosep]
    \item without instruction and without examples
    \item with general task instruction but no examples
    \item with one example per category
    \item with two representative examples per category
    \item with instruction plus one example per category
    \item with instruction plus two examples per category
\end{enumerate}

For the few-shot settings, one or two representative examples from the training set were selected for each of the eight categories and incorporated into the prompt. The LLMs were instructed to output the category index (1–8) for each explanation. We evaluate both classification accuracy and the distributional alignment between LLM predictions and the gold human label distributions, as reported in Table~\ref{tab:classification_results}.

\begin{table*}[h]
\centering
\resizebox{\textwidth}{!}{%
\begin{tabular}{lcccccccc}
\toprule
\multicolumn{1}{l|}{\multirow{2}{*}{\textbf{Classifiers}}} &
  \multicolumn{1}{l|}{\multirow{2}{*}{\textbf{Accuracy}}} &
  \multicolumn{2}{c|}{\textbf{Precision}} &
  \multicolumn{2}{c|}{\textbf{Recall}} &
  \multicolumn{2}{c|}{\textbf{F1}} &
  \multirow{2}{*}{\textit{\textbf{Invalid predictions}}} \\
  \multicolumn{1}{l|}{} &
  \multicolumn{1}{l|}{} &
  macro &
  \multicolumn{1}{l|}{\textit{weighted}} &
  macro &
  \multicolumn{1}{l|}{\textit{weighted}} &
  macro &
  \multicolumn{1}{l|}{\textit{weighted}} &
   \\ \midrule
\textbf{Llama-3.2-3B-Instruct}                     & 0.357 & 0.440     & 0.581        & 0.373     & 0.357        & 0.291     & 0.310        & 0 (0.00\%) \\
+ instruction                                      & 0.229 & 0.379     & 0.465        & 0.281     & 0.229        & 0.227     & 0.256        &918 (29.54\%)\\
+ one example per category                         & 0.340 & 0.393     & 0.540        & 0.343     & 0.340        & 0.255     & 0.293        & 23 (0.74\%)\\
+ two example per category                         & 0.160 & 0.243     & 0.302        & 0.252     & 0.160        & 0.139     & 0.163        &277 (8.91\%)\\
+  instruction + one example per category          & 0.357 & 0.440     & 0.581        & 0.272     & 0.357        & 0.291     & 0.310        & 0 (0.00\%) \\
+  instruction + two example per category          & 0.538 & 0.484     & 0.591        & 0.402     & 0.538        & 0.397     & 0.522        & 0 (0.00\%) \\ \midrule
\textbf{gpt-3.5-turbo}                             & 0.289 & 0.264     & 0.351        & 0.286     & 0.289        & 0.239     & 0.279        & 0 (0.00\%) \\
+ instruction                                      & 0.366 & 0.314     & 0.431        & 0.357     & 0.366        & 0.295     & 0.336        & 0 (0.00\%) \\
+ one example per category                         & 0.175 & 0.162     & 0.244        & 0.155     & 0.175        & 0.139     & 0.182        & 28 (0.90\%)\\
+ two example per category                         & 0.297 & 0.281     & 0.403        & 0.265     & 0.297        & 0.237     & 0.308        & 1 (0.03\%) \\
+  instruction + one example per category          & 0.274 & 0.286     & 0.393        & 0.264     & 0.274        & 0.236     & 0.290        & 36 (1.16\%)\\
+  instruction + two example per category          & 0.305 & 0.317     & 0.420        & 0.301     & 0.305        & 0.262     & 0.303        & 8 (0.26\%) \\ \midrule
\textbf{gpt-4o}                                    & 0.433 & 0.402     & 0.495        & 0.409     & 0.433        & 0.321     & 0.411        & 0 (0.00\%) \\
\textit{+ instruction}                             & 0.410 & 0.465     & 0.536        & 0.438     & 0.410        & 0.357     & 0.404        & 0 (0.00\%) \\
\textit{+ one example per category}                &\textbf{0.594} & 0.530     & 0.619        & 0.486     & \textbf{0.594}       & 0.476     & \textbf{0.583}        & 0 (0.00\%) \\
\textit{+ two example per category}                & 0.589 & \textbf{0.545}     & 0.631        & 0.532     & 0.589        & 0.491     & 0.579        & 0 (0.00\%) \\
\textit{+  instruction + one example per category} & 0.583 & 0.550     & 0.643        & 0.548     & 0.583        & 0.491     & 0.578        & 0 (0.00\%) \\
\textit{+  instruction + two example per category} & 0.574 & 0.541     &\textbf{0.648}        & 0.552     & 0.574        & \textbf{0.492}     & 0.573        & 0 (0.00\%) \\ \midrule
\textbf{DeepSeek-v3}                               & 0.340 & 0.306     & 0.409        & 0.389     & 0.340        & 0.268     & 0.312        & 1 (0.03\%) \\
+ instruction                                      & 0.422 & 0.423     & 0.508        & 0.480     & 0.422        & 0.369     & 0.388        & 0 (0.00\%) \\
+ one example per category                         & 0.540 & 0.483     & 0.592        & 0.514     & 0.540        & 0.461     & 0.529        & 0 (0.00\%) \\
+ two example per category                         & 0.560 & 0.498     & 0.611        & 0.520     & 0.560        & 0.475     & 0.552        & 0 (0.00\%) \\
+  instruction + one example per category          & 0.495 & 0.504     & 0.603        & 0.544     & 0.495        & 0.453     & 0.474        & 0 (0.00\%) \\
+  instruction + two example per category          & 0.526 & 0.519     & 0.626        & \textbf{0.563}     & 0.526        & 0.478     & 0.515        & 0 (0.00\%) \\ \bottomrule
\end{tabular}%
}
\caption{LLM as classifiers results.}
\label{tab:classification_results}
\end{table*}

From the results, we observe that GPT-4o consistently delivers the strongest performance across most experimental configurations, achieving its highest accuracy of 0.594 in the \textit{+ one example per category} setting. Its macro-F1 reaches 0.492 in the \textit{+ instruction + one example per category} setting, while its weighted-F1 peaks at 0.583 in the \textit{+ one example per category} setting, both surpassing all other LLMs. To gain deeper insight, we further analyze the confusion cases under the {+ one example per category} setting, focusing on GPT-4o as a representative model.

\begin{figure}[H]
  \includegraphics[width=\columnwidth]{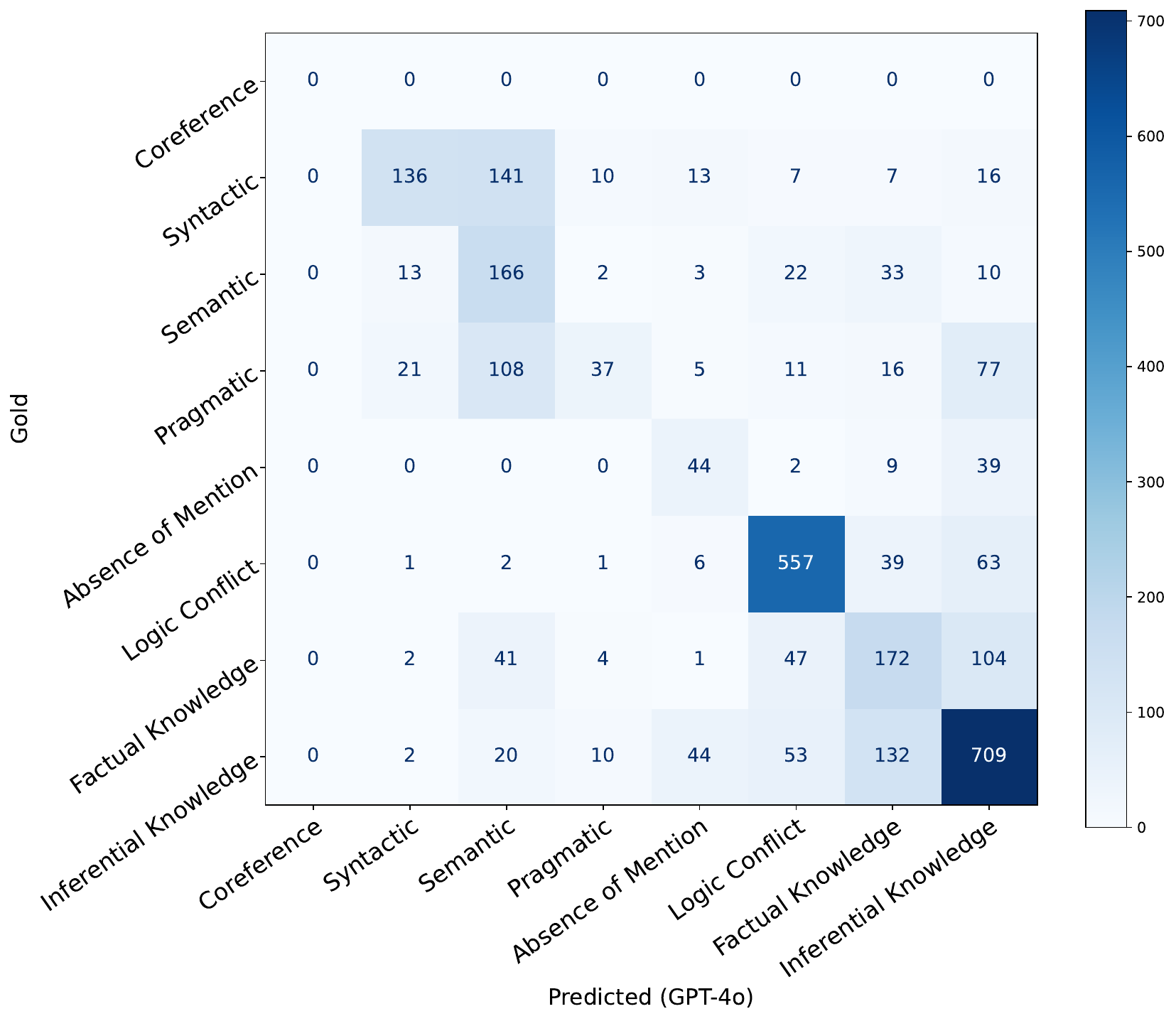}
  \caption{Confusion matrix of GPT-4o on the NLI explanation classification task using the \textit{+ one example per category} prompting style.}
  \label{fig:confusion_classification}
\end{figure}
Figure~\ref{fig:confusion_classification} provides a detailed view of the error patterns in GPT-4o's predictions. We observe that \textit{Syntactic} and \textit{Semantic} are frequently confused, 
indicating that the model has difficulty capturing the fine-grained distinction between structural and meaning-oriented reasoning. 
Similarly, a considerable number of \textit{Factual Knowledge} instances are mislabeled as \textit{Inferential Knowledge}, 
suggesting that GPT-4o often fails to separate lexical associations from broader factual inferences. To further illustrate these confusions, consider the following examples: \\ \\
\colorbox{gray!10}{\textit{Syntactic vs. Semantic}} \\
\textbf{Premise:} A man in a black shirt overlooking bike maintenance. \\
\textbf{Hypothesis:} A man watches bike repairs. \\
\textbf{Explanation:} A man is watching the bike maintenance which is repairs. \\
\textbf{Category (Human):} Syntactic \\
\textbf{Category (GPT-4o):} Semantic \\
\textbf{Analysis:} Human annotators classify this case as \textit{Syntactic}, since the paraphrase between ``maintenance'' and ``repairs'' is treated as a surface-level syntactic variation. 
In contrast, GPT-4o labels it as \textit{Semantic}, interpreting the paraphrase primarily as a meaning equivalence rather than a structural rewording. 
\\ \\
\colorbox{gray!10}{\textit{Inferential vs. Factual Knowledge}} \\
\textbf{Premise:} A blond-haired doctor and her African American assistant looking through new medical manuals. \\
\textbf{Hypothesis:} A doctor is studying. \\
\textbf{Explanation:} Answer: Just because the doctor is studying it doesn't mean he is reading medical manuals. \\
\textbf{Category (Human):} Inferential Knowledge \\
\textbf{Category (GPT-4o):} Factual Knowledge \\
\textbf{Analysis:} Human annotators label this case as \textit{Inferential Knowledge}, since the reasoning requires recognizing the pragmatic gap between ``studying'' in general and ``studying medical manuals'' in particular. 
GPT-4o, however, classifies it as \textit{Factual Knowledge}, suggesting that it grounds the judgment in the surface facts of the premise rather than modeling the inference beyond what is explicitly mentioned. \\

In contrast, \textit{Absence of Mention} is classified with high reliability, as reflected by the strong diagonal concentration in its row. These observations highlight that GPT-4o is more robust when reasoning relies on explicit absence cues, 
while it struggles when categories require distinguishing subtle linguistic or knowledge-based inferences.

To further assess the impact of supervised adaptation, we finetune Llama-3.2-3B-Instruct using LoRA \cite{hu2021loralowrankadaptationlarge}, a parameter-efficient fine-tuning method. We adopt a 50/50 train-test split based on pairID. Fine-tuning is conducted using SFTTrainer with standard causal language modeling objectives and a maximum input length of 512 tokens. The LoRA configuration used is displayed in Table~\ref{tab:hyperparameter_llama}. 

\begin{table}[H]
\centering
\small
\begin{tabular}{ll}
\toprule
\textbf{Hyperparameter} & \textbf{Value}  \\ \midrule
Model     &  Llama-3.2-3B-Instruct       \\
Gradient Accumulation & 4 \\
Max Sequence Length & 512 \\
Warmup Steps            & 50           \\
Scheduler & Cosine \\
Learning Rate           & 2e-4         \\
Batch Size              & 4             \\
Num Epoch               & 3             \\ 
Trainer & SFTTrainer (TRL) \\ \bottomrule
\end{tabular}
\caption{Training hyperparameters used for LoRA fine-tuning on Llama-3.2-3B. LoRA settings: $r$ = 8, $\alpha$ = 16, dropout = 0.05.}
\label{tab:hyperparameter_llama}
\end{table}

The fine-tuned Llama-3.2-3B model achieves an accuracy of 0.509 and a macro-F1 score of 0.302 on the test set. Detailed per-category results are presented in Table~\ref{tab:lora-results}.

\begin{table}[H]
\centering
\resizebox{\columnwidth}{!}{%
\begin{tabular}{lccc}
\toprule
\textbf{Explanation Category} & \textbf{Precision} & \textbf{Recall} & \textbf{F1} \\
\midrule
Coreference            & 0.429 & 0.052 & 0.092 \\
Semantic          & 0.250 & 0.489 & 0.331 \\
Syntactic       & 0.548 & 0.182 & 0.273 \\
Pragmatic         & 0.273 & 0.200 & 0.231 \\
Absence of Mention                & 0.000 & 0.000 & 0.000 \\
Logic Conflict        & 0.735 & 0.758 & 0.746 \\
Factual Knowledge                 & 0.138 & 0.041 & 0.064 \\
Inferential Knowledge & 0.562 & 0.861 & 0.680 \\
\multicolumn{4}{c}{\cellcolor[HTML]{EFEFEF}\textit{\textbf{Summary}}} \\
accuracy      & \multicolumn{3}{c}{0.509} \\
F1 Score (macro)      & \multicolumn{3}{c}{0.302} \\
\bottomrule
\end{tabular}
}
\caption{LoRA fine-tuning results using Llama-3.2-3B-Instruct on the explanation categorization task.}
\label{tab:lora-results}
\end{table}

While zero-shot prompting offers a lightweight baseline, these results suggest that parameter-efficient fine-tuning can boost performance in structured reasoning categories such as \textit{Logical Conflict} and \textit{Inferential Knowledge}. However, performance remains limited in categories such as \textit{Factual Knowledge}, which require external world knowledge, and \textit{Absence of Mention}, where low performance may be attributed to the small number of training examples.

We accessed GPT-3.5 and GPT-4o via OpenAI's hosted API and DeepSeek-V3 via DeepSeek's hosted API. Experiments with Llama-3.2-3B-Instruct were run on a single NVIDIA A100 GPU.

\section{Human Highlight IAA}
\label{sec:highlight-iaa}

To understand whether human-generated highlights are consistent and reproducible, we conducted a highlight-level inter-annotator agreement (IAA) study on 201 items from the e-SNLI dataset. Two annotators were asked to highlight the parts of the premise and hypothesis that support the given explanation. Each item included the premise, hypothesis, gold label and the explanation.

We measured agreement using Intersection over Union (IoU). The results are as follows:

\begin{itemize}
\item \textbf{Annotator 1 vs Annotator 2}: 0.889
\item \textbf{Annotator 1 vs e-SNLI Highlight}: 0.659
\item \textbf{Annotator 2 vs e-SNLI Highlight}: 0.712
\end{itemize}

These results show that the two annotators had high agreement with each other, suggesting that the highlighting task is fairly consistent when done by different people. However, their agreement with the original e-SNLI highlights is lower, which means there are some differences in how people choose text spans, even when they agree on the explanation. This may be partially attributed to differences in annotation setup: in e-SNLI, the same annotator provided the NLI label, explanation, and highlight jointly, whereas in our IAA study, annotators re-annotated highlights for a given explanation under fixed label and span constraints. Although we adopted the same span-level constraints as e-SNLI (e.g., highlighting only the hypothesis for neutral items), our task required linking highlights to prewritten explanations rather than authoring them jointly, introducing a structural difference that may affect highlight choices.

\section{Prompting Templates for Generating Model Explanations}
\label{sec:appx-prompt-generation}

For the generation experiments, we prompt three LLMs to generate NLI explanations: GPT-4o, DeepSeek-V3, and Llama-3.3-70B-Instruct. We accessed GPT-4o via OpenAI's hosted API and DeepSeek-V3 via DeepSeek's hosted API. The generation experiments using Llama-3.3-70B-Instruct were conducted on two NVIDIA A100 GPUs.

Table~\ref{tab:generation-prompt} presents the prompt templates used to generate NLI explanations from LLMs. These templates are adapted and refined based on the approach of \citet{chen2024seeingbigsmallllms}. For LLMs that imply a “system” role within their chat format, the “system” role content is unset to maintain alignment with the design choices applied to other LLMs.

\begin{table*}[h]
\centering
\resizebox{0.9\textwidth}{!}{
\footnotesize
\begin{tabularx}{\textwidth}{@{}l>{\raggedright\arraybackslash}X@{}}
\toprule
\textbf{Mode} & \textbf{General Instruction Prompt} \\ \midrule
\textit{baseline} &

You are an expert in Natural Language Inference (NLI). Please list all possible explanations for why the following statement is \{gold\_label\} given the content below without introductory phrases.  

Context: \{premise\}, Statement: \{hypothesis\}  \\[2ex] \midrule
\textit{highlight indexed} &

You are an expert in Natural Language Inference (NLI). Your task is to generate possible explanations for why the following statement is \{gold\_label\}, focusing on the highlighted parts of the sentences. 

Context: \{premise\}, Highlighted word indices in Context: \{highlighted\_1\}  

Statement: \{hypothesis\}, Highlighted word indices in Statement: \{highlighted\_2\}  

Please list all possible explanations without introductory phrases. \\[2ex] \midrule

\textit{highlight in-text} &

You are an expert in Natural Language Inference (NLI). Your task is to generate possible explanations for why the following statement is \{gold\_label\}, focusing on the highlighted parts of the sentences. Highlighted parts are marked in ``**''.  

Context: \{marked\_premise\}  Statement: \{marked\_hypothesis\}  

Please list all possible explanations without introductory phrases.  \\[2ex] 

\midrule
\textit{highlight generation} &

You are an expert in NLI. Based on the label '{gold\_label}', highlight relevant word indices in the premise and hypothesis. Highlighting rules:
\- For entailment: highlight at least one word in the premise.
\- For contradiction: highlight at least one word in both the premise and the hypothesis.
\- For neutral: highlight only in the hypothesis.

Premise: \{premise\}, Hypothesis: \{hypothesis\}, Label: \{gold\_label\}

Please list **3** possible highlights using word index in the sentence without introductory phrases. Answer using word indices **starting from 0** and include punctuation marks as tokens (count them). Respond strictly this format:
        
Highlight 1: 

Premise\_Highlighted: [Your chosen index(es) here] 

Hypothesis\_Highlighted: [Your chosen index(es) here]

Highlight 2:
...
\\[2ex] 

\midrule
\textit{taxonomy (two-stage)} &

You are an expert in Natural Language Inference (NLI). Given the following taxonomy with description and one example, generate as many possible explanations as you can that specifically match the reasoning type described below. The explanation is for why the following statement is \{gold\_label\}, given the content.  

The explanation category for generation is:  
\{taxonomy\_idx\}: \{description\}  

Here is an example:  
Premise: \{few\_shot{[}'premise'{]}\}, Hypothesis: \{few\_shot{[}'hypothesis'{]}\}  

Label: \{few\_shot{[}'gold\_label'{]}\}, Explanation: \{few\_shot{[}'explanation'{]}\}

Now, consider the following premise and hypothesis: 

Context: \{premise\}  
Statement: \{hypothesis\}

Please list all possible explanations for the given category without introductory phrases.\\[2ex]

\midrule
\textit{taxonomy end-to-end} &

You are an expert in Natural Language Inference (NLI). Your task is to examine the relationship between the following content and statement under the given gold label, and:
First, identify all categories for explanations from the list below (you may choose more than one) that could reasonably support the label.
Second, for each selected category, generate all possible explanations that reflect that type.

The explanation categories are:

\{taxonomy\_idx\}: \{description\}  

Context: \{premise\}, Statement: \{hypothesis\}, Label: \{gold\_label\}

Please list all possible explanations without introductory phrases for all the chosen categories. 

Start directly with the category number and explanation, following the strict format below:

1. Coreference: - [Your explanation(s) here]  

... (continue for all reasonable categories)\\[2ex]

\midrule
\textit{taxonomy two-stage (classification)} & 

You are an expert in Natural Language Inference (NLI). Your task is to identify all applicable reasoning categories for explanations from the list below that could reasonably support the label. Please choose at least one category and multiple categories may apply. One example for each category is listed as below:

\{examples\_text\}

Given the following premise and hypothesis, identify the applicable explanation categories:

Premise: \{premise\}, Hypothesis: \{hypothesis\}, Label: \{gold\_label\}

Respond only with the numbers corresponding to the applicable categories, separated by commas, and no additional explanation.\\[2ex]
\bottomrule
\end{tabularx}}
\caption{Instruction prompts for LLMs to generate NLI explanations (all prompts are issued as user messages in the chat format).}
\label{tab:generation-prompt}
\end{table*}

\FloatBarrier

\section{Additional Generation Results} \label{sec:additional_results}

Table~\ref{tab:additional_results} presents the full evaluation results of our explanation generation experiments, covering two highlight formats (indexed vs. in-text) and both human-provided and model-generated highlights.

\begin{table*}[ht]
\centering
\resizebox{\textwidth}{!}{%
\begin{tabular}{@{}lccccccccccc@{}}
\toprule
\multirow{2}{*}{\textbf{Mode}} &
  \multirow{2}{*}{\textbf{Cosine}} &
  \multirow{2}{*}{\textbf{Euclidean}} &
  \multicolumn{2}{c}{\textbf{\begin{tabular}[c]{@{}c@{}}1gram\end{tabular}}} &
  \multicolumn{2}{c}{\textbf{\begin{tabular}[c]{@{}c@{}}2gram\end{tabular}}} &
  \multicolumn{2}{c}{\textbf{\begin{tabular}[c]{@{}c@{}}3gram\end{tabular}}} &
  \multirow{2}{*}{\textbf{BLEU}} &
  \multirow{2}{*}{\textbf{ROUGE-L}} &
  \multirow{2}{*}{\textbf{Avg\_len}} \\ 
  \cmidrule(lr){4-5} \cmidrule(lr){6-7} \cmidrule(lr){8-9}
  &
  &
  &
  \textbf{Word} &
  \textbf{POS} &
  \textbf{Word} &
  \textbf{POS} &
  \textbf{Word} &
  \textbf{POS} &
  &
  &
  \\ \midrule
GPT4o \\
\textit{\, \, baseline}                
& 0.556 & 0.524 & 0.291 & 0.882  & 0.117 & 0.488 & 0.049 & 0.226 & 0.051 & 0.272 & 24.995 \\ 
\textit{\, \, human highlight (indexed)} & 0.549 & 0.521 & 0.395 & 0.882 & 0.116 & 0.478 & 0.050 & 0.219 & 0.047 & 0.264 & 30.771 \\  
\textit{\, \, human highlight (in-text)} & 0.519 & 0.511 & 0.367 & 0.873 & 0.085 & 0.442 & 0.031 & 0.187 & 0.034 & 0.269 & 28.606 \\
\textit{\, \, model highlight (indexed)} 
& 0.554 & 0.522 & 0.402 & 0.878 & 0.124 & 0.481 & 0.053 & 0.222 & 0.051 & 0.269 & 28.240 \\
\textit{\, \, model highlight (in-text)}
& 0.555 & 0.523 & 0.380 & 0.888 & 0.109 & 0.468 & 0.044 & 0.208 & 0.044 & 0.270 & 28.160 \\  
\textit{\, \, model taxonomy (two-stage)}  
& 0.593 & 0.537 & 0.418 & 0.886 & 0.128 & 0.495 & 0.071 & 0.242 & 0.071 & 0.314 & 19.991 \\
\textit{\, \, model taxonomy (end-to-end)} 
& \textbf{0.608} & \textbf{0.540} & \textbf{0.437} & \textbf{0.898} & \textbf{0.166} & \textbf{0.511} & \textbf{0.083} & \textbf{0.255} & \textbf{0.074} & \textbf{0.323} & 26.672 \\ \midrule
DeepSeek-v3 \\
\textit{\, \, baseline}           
& 0.428 & 0.490 & 0.369 & 0.847 & 0.087 & 0.449 & 0.034 & 0.195 & 0.042 & 0.245 & 20.288 \\
\textit{\, \, human highlight (indexed)}  & 0.463 & 0.498 & 0.358 & 0.864 & 0.084 & 0.436 & 0.033 & 0.184 & 0.035 & 0.243 & 29.293 \\
\textit{\, \, human highlight (in-text)} & 0.551 & 0.522 & 0.362 & 0.885 & 0.091 & 0.449 & 0.033 & 0.191 & 0.036 & 0.261 & 28.527 \\
\textit{\, \,  model highlight (indexed)}  & 0.464 & 0.499 & 0.364 & 0.861 & 0.091 & 0.450 & 0.037 & 0.196 & 0.034 & 0.242 & 27.301 \\
\textit{\, \, model highlight (in-text)}  & 0.447 & 0.457 & 0.341 & 0.869 & 0.073 & 0.422 & 0.026 & 0.171 & 0.030 & 0.248 & 31.328 \\
\textit{\, \, model taxonomy (two stage)}          & 0.544 & 0.522 & 0.391 & 0.884 & 0.122 & 0.475 & 0.055 & 0.219 & 0.057 & 0.293 & 20.894 \\
\textit{\, \, model taxonomy (end-to-end)} & \textbf{0.556} & \textbf{0.528} & \textbf{0.404} & \textbf{0.897} & \textbf{0.140} & \textbf{0.486} & \textbf{0.067} & \textbf{0.233} & \textbf{0.063} & \textbf{0.306} & 25.960 \\ \midrule
Llama-3.3-70B \\
\textit{\, \, baseline} & 0.466 & 0.496 & 0.392 & 0.863 & 0.106 & 0.478 & 0.044 & 0.224 & 0.046 & 0.250 & 27.148 \\
\textit{\, \, human highlight (indexed)} & 0.453 & 0.484 & 0.362 & 0.859 & 0.082 & 0.446 & 0.031 & 0.194 & 0.035 & 0.228 & 29.912 \\
\textit{\, \, human highlight (in-text)}& 0.499 & 0.505 & 0.348 & 0.875 & 0.059 & 0.415 & 0.019 & 0.165 & 0.024 & 0.270 & 34.827 \\
\textit{\, \, model highlight (indexed)} & 0.367 & 0.478 & 0.317 & 0.807 & 0.065 & 0.408 & 0.024 & 0.173 & 0.031 & 0.199 & 24.987 \\
\textit{\, \, model highlight (in-text)}& 0.400 & 0.486 & 0.300 & 0.831 & 0.047 & 0.385 & 0.014 & 0.150 & 0.021 & 0.227 & 29.763 \\
\textit{\, \, model taxonomy (two-stage)}  & \textbf{0.609} & \textbf{0.541} & \textbf{0.444} & 0.889 & \textbf{0.167} & \textbf{0.512} & \textbf{0.082} & \textbf{0.256} & \textbf{0.078} & 0.321 & 22.340 \\
\textit{\, \, model taxonomy (end-to-end)} & 0.505 & 0.510 & 0.383 & \textbf{0.896} & 0.110 & 0.499 & 0.048 & 0.232 & 0.047 & 0.262 & 28.870 \\
\midrule
\end{tabular}%
}
\caption{Full evaluation results for LLM-generated explanations (lexical, morphosyntactic, semantic, and summarization levels).}
\label{tab:additional_results}
\end{table*}

\paragraph{Human Highlights vs. Model Generated Highlights} Overall, model highlights achieve comparable performance to human highlights across most lexical and semantic metrics, with slight improvements in certain surface-level features (e.g., BLEU, ROUGE-L). However, these gains are often marginal. Notably, models like Llama-3.3-70B show a larger drop in similarity metrics when using model-generated highlights, indicating that automatic highlight classification may not always align with human judgment.

\paragraph{Indexed vs. In-text}
We compare the indexed and in-text variants of human and model highlights to assess whether highlight format affects similarity scores. Across all three models, the performance differences between the two formats are generally minor with the indexed variant performing slightly better. For instance, GPT-4o yields similar scores in both settings (e.g., cosine: 0.549 vs. 0.519 for human highlights; 0.554 vs. 0.555 for model highlights). The same trend holds for DeepSeek-v3 and Llama-3.3-70B, where average performance differences across metrics remain negligible. 

\section{Human Validation of Model-Generated Explanations}\label{sec:human-validation}

Table~\ref{tab:taxonomy_validation} reports human validation results for model-generated explanations, broken down by taxonomy category. This analysis helps us better examine how explanation faithfulness and taxonomy alignment vary across different reasoning types.

\begin{table}[H]
\centering
\small
\resizebox{\columnwidth}{!}{%
\begin{tabular}{lcccr}
\toprule
\textbf{Taxonomy} & \textbf{Q1 Yes (\%)} & \textbf{Q1 No (\%)} & \textbf{Q2 Yes (\%)} & \textbf{Q2 No (\%)} \\
\midrule
Coreference               & 269 (97.46)  & 7 (2.54)    & 158 (57.25)  & 118 (42.75) \\
Syntactic                 & 780 (99.87)  & 1 (0.13)    & 741 (94.88)  & 40 (5.12)   \\
Semantic                  & 1716 (95.12) & 88 (4.88)   & 1273 (70.57) & 531 (29.43) \\
Pragmatic                 & 131 (99.24)  & 1 (0.76)    & 109 (82.58)  & 23 (17.42)  \\
Absence of Mention        & 3794 (99.16) & 32 (0.84)   & 3538 (92.47) & 288 (7.53)  \\
Logic Conflict            & 428 (98.85)  & 5 (1.15)    & 273 (63.05)  & 160 (36.95) \\
Factual Knowledge         & 949 (99.16)  & 8 (0.84)    & 789 (82.45)  & 168 (17.55) \\
Inferential Knowledge     & 161 (98.17)  & 3 (1.83)    & 139 (84.76)  & 25 (15.24)  \\
\bottomrule
\end{tabular}%
}
\caption{Human validation results for model-generated explanations by taxonomy category. Q1: Whether the explanation supports the gold label. Q2: Whether the explanation matches the assigned taxonomy.}
\label{tab:taxonomy_validation}
\end{table}

Across all categories, validation question 1 — evaluating whether the explanation supports the gold NLI label — yields consistently high agreement, with most categories exceeding 98\% ``Yes'' responses. This indicates that the generated explanations are largely faithful to the NLI decision, regardless of the reasoning type. In contrast, validation question 2 — assessing whether the explanation aligns with the specified taxonomy — shows greater variation across categories. Categories such as \textit{Syntactic} and \textit{Absence of Mention} achieve the highest taxonomy agreement, with 94.88\% and 92.47\% of explanations remaining consistent with their respective reasoning types. These categories tend to involve explicit cues, which may be easier for LLMs to identify and replicate during generation. For example, explanations like ``A is a rephrase of B'' or ``A in the premise is rephrased in the hypothesis'' are common and prototypical forms of the \textit{Syntactic} category. Similarly, in the \textit{Absence of Mention} category, model outputs often include patterns such as ``The premise discusses A but does not mention B'' or ``A is absent from the premise'', which directly map onto the intended reasoning structure and are relatively easy to pattern-match.

In contrast, categories like \textit{Coreference} (57.25\%) and \textit{Logic Conflict} (63.05\%) show significantly lower alignment with the taxonomy categories. These types require discourse-level understanding or implicit logical inference, such as tracking entity references across clauses or identifying contradictions in different logical forms (temporal contradiction, location contradiction, gender conflict, etc.). Such reasoning is more abstract and difficult to control through prompting, which likely explains the increased rate of taxonomy mismatches.

Categories such as \textit{Semantic}, \textit{Factual Knowledge}, and \textit{Inferential Knowledge} fall in an intermediate range (70–85\%), likely due to their broader and more flexible definitions. For instance, semantic reasoning can often overlap with world knowledge or pragmatic cues, making it harder for models (and annotators) to sharply distinguish the boundaries of the category. This pattern is consistent with our IAA findings reported in §\ref{subsec:iaa}, where we observed lower precision for \textit{Semantic} (0.643) and lower recall for \textit{Factual Knowledge} (0.652). These results point to potential ambiguities in distinguishing these categories from others, particularly from \textit{Inferential Knowledge}.

\end{document}